\documentclass{article}

\usepackage{arxiv}

\usepackage[utf8]{inputenc} 
\usepackage[T1]{fontenc}    
\usepackage{hyperref}       
\usepackage{url}            
\usepackage{booktabs}       
\usepackage{amsfonts}       
\usepackage{nicefrac}       
\usepackage{microtype}      
\usepackage{lipsum}		
\usepackage{graphicx}

\usepackage{caption}
\usepackage{subcaption}

\usepackage[numbers]{natbib}
\usepackage{doi}
\usepackage{xcolor}
\usepackage{multirow}
\usepackage{longtable}
\usepackage{tabularx}
\usepackage{tabu}
\usepackage{amsmath}
\usepackage{lscape}

\title{Sample Selection Bias in Machine Learning for Healthcare}

\author{{\hspace{1mm}Vinod Kumar Chauhan$^{1,2}$\thanks{Corresponding author: Vinod Kumar Chauhan (\href{mailto:vinod.kumar@eng.ox.ac.uk}{vinod.kumar@eng.ox.ac.uk})},\; Lei Clifton$^{3}$,\; Achille Sala\"un$^1$,\; Huiqi Yvonne Lu$^1$,\; Kim Branson$^4$,}
\And
Patrick Schwab$^4$,\; Gaurav Nigam$^{5}$,\; David A. Clifton$^{1,6}$
\And
	{}\\$^1$Department of Engineering Science,
	University of Oxford, UK\\
        $^2$Department of Computer and Information Sciences, University of Strathclyde\\
        $^3$Nuffield Department of Primary Care Health Sciences, University of Oxford, UK\\
        $^4$Biomedical AI Group, GSK\\
        $^5$Nuffield Department of Medicine, University of Oxford, Oxford, UK\\
        $^6$Oxford-Suzhou Institute of Advanced Research (OSCAR), Suzhou, China
}

\renewcommand{\undertitle}{Accepted to ACM Transactions on Computing for Healthcare
\url{https://doi.org/10.1145/3761822}}
\renewcommand{\shorttitle}{Sample Selection Bias in Machine Learning for Healthcare}

\hypersetup{
    pdftitle={Sample Selection Bias in Machine Learning for Healthcare},
    pdfsubject={cs.LG, cs.AI},
    pdfauthor={V. K. Chauhan},
    pdfkeywords={Machine Learning, Sample Selection Bias, Healthcare},
}

\begin{document}
\maketitle

\begin{abstract}
    While machine learning algorithms hold promise for personalised medicine, their clinical adoption remains limited, partly due to biases that can compromise the reliability of predictions. In this paper, we focus on sample selection bias (SSB), a specific type of bias where the study population is less representative of the target population, leading to biased and potentially harmful decisions.
    Despite being well-known in the literature, SSB remains scarcely studied in machine learning for healthcare. Moreover, the existing machine learning techniques try to correct the bias mostly by balancing distributions between the study and the target populations, which may result in a loss of predictive performance.
    To address these problems, our study illustrates the potential risks associated with SSB by examining SSB's impact on the performance of machine learning algorithms.
    Most importantly, we propose a new research direction for addressing SSB, based on the target population identification rather than the bias correction.
    Specifically, we propose two independent networks (T-Net) and a multitasking network (MT-Net) for addressing SSB, where one network/task identifies the target subpopulation which is representative of the study population and the second makes predictions for the identified subpopulation.
    Our empirical results with synthetic and semi-synthetic datasets highlight that SSB can lead to a large drop in the performance of an algorithm for the target population as compared with the study population, as well as a substantial difference in the performance for the target subpopulations that are representative of the selected and the non-selected patients from the study population. Furthermore, our proposed techniques demonstrate robustness across various settings, including different dataset sizes, event rates, and selection rates, outperforming the existing bias correction techniques.
\end{abstract}


\section{Introduction}
\label{sec_intro}
Machine learning algorithms have demonstrated high diagnostic and prognostic accuracy in identifying and categorising diseases, promising personalised interventions and informed decision-making in healthcare \cite{rajkomar2019machine,topol2019high}. Their ability to analyse vast datasets and uncover hidden patterns surpasses traditional techniques and sometimes even human experts in task-specific applications, such as image-recognition tasks in radiology \cite{jumper2021highly,hosny2018artificial}. However, despite their compelling potential and an increasing number of studies every year, their clinical adoption remains constrained by various challenges \cite{vokinger2021mitigating}, including biases that can undermine the validity of machine learning models. Among these, sample selection bias (SSB) has received limited attention in machine learning for healthcare, despite being recognised as a fundamental pitfall in the research design of clinical studies \cite{yu2020one}, posing a significant hurdle to real-world applications.

\begin{figure}[htb!]
    \centering
    \includegraphics[width=0.85\linewidth]{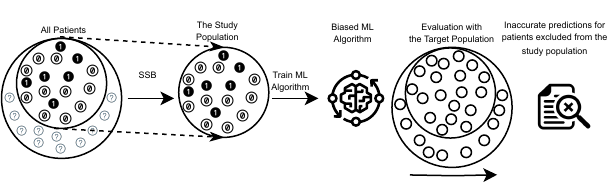}
    \caption{Illustration of SSB and its impact: We start with all patients who participated in the study, where we have known outcomes for patients inside the inner circle (denoted using 0 and 1) while the rest have unknown outcomes (denoted using `?'). The study population includes patients only from the inner circle, i.e., those patients who have known outcomes, resulting in SSB. A machine learning (ML) algorithm trained on such a dataset does not have information about the non-selected patients. When such an algorithm is deployed and evaluated with the target population which contains patients that are representative of the selected and the non-selected from the study population, then it may lead to inaccurate predictions and harmful decisions for the subpopulations representative of the non-selected patients.}
    \label{fig_SSB}
\end{figure}

Inherent in clinical studies, which serve as a source of data for machine learning, is the practice of sample selection, characterised by strict criteria for the inclusion and exclusion of patients \cite{parsons2023independent}. However, these studies don’t add any bias to results as long as the study population is representative of the target population. The study population is defined as data used to develop the model and generally split into training, (internal) validation and test datasets; and the target population is defined as the independent external data, i.e., deployment data or external validation data, respectively. SSB results from a non-uniform sampling process to select patients in the study population, leading to differences in the distributions of the study population and the target population \cite{porta2014dictionary,bonita2006basic,zhelonkin2016robust,boonstra2021simulation}. For example, suppose in a clinical study, the outcomes of some patients are not available due to a loss to follow-up or competing events, and considering only those patients who completed the study introduces SSB in the dataset \cite{banack2019investigating}. When machine learning algorithms are trained on such a biased dataset and employed on an unbiased dataset, it might lead to inaccurate predictions and suboptimal care for patient subpopulations not represented in the study population \cite{rojas2023selection,Ellen2024}. Another example is the self-nomination of patients for a clinical study, which may not represent the entire population, such as the UK Biobank and might lead to SSB \cite{swanson2012uk,bradley2022addressing}. Fig.~\ref{fig_SSB} provides an example illustrating SSB and its impact on a prediction algorithm's performance.

SSB is well known for decades \cite{gronau1974wage,heckman1990varieties,berk1983introduction}, and it has been discussed across several disciplines, including social science and econometrics \cite{heckman1976common,heckman1979sample}, environmental studies \cite{wagenaar2021improved}, finance \cite{Banasik2003}, causality \cite{hernan2004structural,bia2023double}, fairness \cite{Du2022fair}, healthcare \cite{Navarron2021}, and machine learning \cite{zadrozny2004learning}. It is also an important consideration for clinical study design as the results of a biased study may not apply to a real patient population \cite{bonita2006basic,porta2014dictionary}.
Despite being well-known and a threat to the validity of algorithms \cite{christiansen2023device}, SSB has been scarcely studied in machine learning for healthcare \cite{yu2020one}. This could be potentially because (i) often less proportion of patients are excluded from the study population, making it hard to find enough data to correct the bias, (ii) most of the algorithms are not developed for the real-world deployment, (iii) the bias is sometimes hidden under the assumed randomness in selection \cite{salaun2023interpretable}, (iv) risk estimation primarily belongs to causality, a principled approach to decision-making that examines cause-and-effect relationships \cite{Chauhan2025beyond}. Causality has developed formal techniques, including causal diagrams, to address SSB \cite{weuve2012accounting,smith2020selection}, and (v) SSB might not always lead to inaccurate predictions for the non-selected patients, as they might not influence the algorithm's decision-making process.

It is not always possible to avoid SSB in the study but to utilise valuable information hidden in such a biased dataset, it is necessary to address the bias to ensure that decisions are made prudently and with reduced bias. Machine learning for healthcare lacked research to address SSB, however, the machine learning community has developed several techniques to address SSB \cite{zadrozny2004learning} and to address distribution mismatch between the source and the target domains, such as covariate shift \cite{shimodaira2000improving} or domain-adaptation \cite{kouw2019review}, which have a setting similar to SSB. The existing machine learning techniques correct for the bias by aligning distributions of the study population and the target population, which could lead to a loss of predictive performance of the algorithm \cite{zadrozny2004learning,huang2006correcting,kouw2019review}. Moreover, in healthcare, the non-selected patients form a subpopulation that could greatly differ from the study population, making it hard to correct for the bias. Thus, the existing techniques may perform poorly for subpopulations that are representative of the non-selected subpopulation compared to those representative of the selected subpopulation.

To address the above problems, our study presents two-fold contributions. Firstly, we present an empirical study with synthetic and semi-synthetic datasets to highlight the importance of addressing SSB in machine learning for healthcare. Our study demonstrates that SSB leads to a large performance gap for machine learning algorithms between the study and the target populations as well as a substantial gap between the target subpopulations representing the non-selected and the selected patients.
Secondly, we propose a new research direction for addressing SSB, focusing on identifying the target population rather than correcting the bias. This approach to address SSB involves first identifying patients from the target population who are representative of the study population, followed by predictions for the identified patients.
The proposed approach for addressing SSB makes sense in healthcare settings where the non-selected patients (e.g., those lost to follow-up) form a small subpopulation of the study population. This is in contrast to domain adaptation techniques, where the target domain (i.e., the target population) is assumed to differ from the source domain (i.e., the study population) and the domain adaptation algorithms are supposed to work for the target domain.
Further, as concrete implementations of the proposed research direction, we propose two independent networks (T-Net) and a multitasking network (MT-Net) techniques for addressing SSB. T-Net develops two networks, one for identification, i.e., to predict the selection of a patient to the study population, and the other for the underlying prediction problem. Similarly, MT-Net develops a multitasking network for two tasks, one for the identification and the other for the underlying prediction problem. Finally, we analyse the performance of the proposed techniques against the state-of-the-art baselines using synthetic and semi-synthetic datasets to prove their effectiveness.

The contributions of the paper are summarised as given below.
\begin{itemize}
    \item We present an empirical study with synthetic and semi-synthetic datasets to highlight SSB in machine learning for healthcare, which is otherwise ignored but can lead to biased and harmful decisions.
    \item We propose a new research direction to address SSB in machine learning for healthcare, focusing on identification of the target subpopulation representative of the study population rather than the bias correction, and hence avoiding loss of predictive performance due to aligning distributions between the study and the target populations.
    \item We propose T-Net and MT-Net, as concrete techniques under the proposed research direction, for addressing SSB, which have two networks/tasks for the population identification and the underlying prediction task. We present an empirical analysis of the proposed techniques against the state-of-the-art baselines using synthetic and semi-synthetic datasets, which show the robustness of the proposed T-Net and MT-Net under different settings.
\end{itemize}

The rest of the paper is organised as: Section~\ref{sec_problem_def} provides a problem definition of SSB, while Section~\ref{sec_related_work} presents a brief overview of literature related to SSB. The proposed research direction and T-Net and MT-Net are discussed in Section~\ref{sec_methods}, followed by performance analysis with synthetic and semi-synthetic datasets in Section~\ref{sec_results}. Discussion on the significance of results is presented in Section~\ref{sec_discussion}, and finally, the concluding remarks are presented in Section~\ref{sec_conclusion}.

\section{Background}
\label{sec_problem_def}
In SSB, a study population is selected non-randomly from the general population. In contrast, the target population is selected uniformly from the general population as at the time of deployment patients can belong to any subpopulation. Generally, the selected patients have defined outcomes, i.e., labels available along with features but for the non-selected patients labels are missing and that's why they are not included in the study population. For example, a loss to follow-up or survival bias may lead to SSB as patients lacking labels are not included in the study \cite{banack2019investigating}.
A naive approach to developing a machine learning algorithm using a biased dataset could result in a biased algorithm. Such an algorithm may fail to accurately make predictions for the patients representative of the non-selected subpopulation at the time of deployment, as the algorithm lacks information about this subpopulation.
So, the challenge with SSB involves developing predictive algorithms using biased labelled samples of the selected patients and unlabelled samples of the non-selected patients, to develop accurate and useful algorithms for making predictions. While the ideal scenario involves mitigating the bias to ensure fair and representative models, the primary objective remains the development of algorithms that effectively leverage available data to provide accurate predictions, despite the presence of bias.

Suppose $\mathbf{x} \in \mathbf{X}$ is a $d$-dimensional feature vector representing a patient, $y \in \mathbf{y}$ is the corresponding outcome and $s \in \mathbf{s}$ is a random selection variable which takes value $1$ if patient $\mathbf{x}$ is selected in the study population or $0$ otherwise. For the sake of clarity and simplicity, we focus on binary classification, i.e., risk prediction task. However, our work can be extended to other problem settings, such as regression. Let $\mathbf{X}=\mathbf{X}^0 \cup \mathbf{X}^1 \in \mathbb{R}^d$ be a random sample from the general population, where $\mathbf{x \in X^1}$ has corresponding label $y \in \mathbf{y^1}$ and are selected in the study population, i.e., $s=1$, and $\mathbf{x \in X^0}$ does not have access to corresponding label and are not selected into the study population, i.e., $s=0$. We assume access to labelled and unlabelled samples as $\mathbf{D^1}=\{(\mathbf{x}^1_1,s_1=1, y^1_1), (\mathbf{x}^1_2,s_2=1, y^1_2),..., (\mathbf{x}^1_n,s_n=1, y^1_n)\}$ of $n$ data points, and $\mathbf{D^0}=\{(\mathbf{x}^0_1,s_1=0), (\mathbf{x}^0_2,s_2=0),..., (\mathbf{x}^0_m,s_m=0)\}$ of $m$ data points, respectively. If patients are selected non-randomly in the study population then $\mathbf{D^1}$ forms a biased distribution of true distribution $\mathbf{X} \times \mathbf{s} \times \mathbf{y}$, and leads to SSB. Thus, SSB has a distribution mismatch between the study population and the target population. \citet{zadrozny2004learning} first formalised SSB in machine learning and discussed four cases based on the dependence of $s$ on $(\mathbf{x}, y)$, and also aligned these cases to missingness, treating SSB as a missingness handling issue, since the outcome is missing for the non-selected patients.
\begin{enumerate}
    \item[i.] If $s$ is independent of $(\mathbf{x}, y)$ then the study population is representative of the target population, and there is no SSB \cite{salaun2023interpretable}. This is a case of outcomes of the non-selected patients missing-completely-at-random.
    \item[ii.] If $s$ depends only on $\mathbf{x}$, i.e., $P(s|\mathbf{x},y)=P(s|\mathbf{x})$ then SSB exists \cite{liu2023combining}, and this is a case of outcomes of the non-selected patients missing-at-random.
    \item[iii.] If $s$ depends only on $y$, i.e., $P(s|\mathbf{x},y)=P(s|y)$ then SSB exists, and this is a case where prior probabilities of the outcomes change \cite{bishop1995neural}.
    \item[iv.] If there is no independence between $\mathbf{x}$, $y$, and $s$, and the selection depends on an observed set of features $\mathbf{x}_s$, then SSB exists \cite{brewer2021addressing}. This hinders our ability to construct a reliable mapping from features to outcomes unless we can access an additional set of features $\mathbf{x}_s$, ensuring that $P(s|\mathbf{x}_s, \mathbf{x}, y) = P(s|\mathbf{x}_s)$ for all examples, including those with $s = 0$. This is a case of outcomes of the non-selected patients missing-not-at-random.
\end{enumerate}

In our study, we focus on the case (ii) since it is the most important SSB case in practice and it is more realistic to assume the case (ii) than to assume the case (i) \cite{zadrozny2004learning}. Secondly, this is the most widely assumed case, while using multiple imputations to address SSB, in statistical and clinical literature \cite{liu2023combining}.
SSB has connections to other machine learning problems, including missingness and covariate shift or domain adaption. However, SSB is a bit different to standard missingness handling problems where some attributes of a patient are missing while other patients from the same population have those attributes \cite{chauhan2024continuous}. SSB is a special case of the missingness problem where the outcome is missing rather than attributes \cite{dost2022selection}, and it may be difficult to impute the missing labels of the non-selected patients as labels of the entire subpopulation are missing.
SSB also intertwines with the broader concept of domain adaptation. Domain adaptation ensures machine learning algorithms trained on a source domain can perform well in a different target domain with varying feature distributions \cite{kouw2019review}. Specifically, covariate shift addresses differences in input feature distributions between source and target domains, i.e., a setting where $P^{\text{source}}(x) \neq P^{\text{target}}(x)$ while everything else remains unchanged. Since for SSB $P(s|\mathbf{x},y)=P(s|\mathbf{x})$ and $P(\mathbf{x}|s=1) \neq P(\mathbf{x}|s)$, this implies $P^{\text{study}}(\mathbf{x}) \neq P^{\text{target}}(\mathbf{x})$. Therefore, SSB and covariate shift share fundamental similarities in addressing mismatches in feature distributions between the study population (source domain) and the target population (target domain). However, they originate from different settings because for SSB the target population is supposed to be a super-set of the study population while for covariate shift the target domain is supposed to be different than the source domain.

\section{Related Work}
\label{sec_related_work}
Here, we discuss work on SSB related to machine learning for healthcare as well as to other fields of machine learning.

\subsection{SSB in Machine Learning for Healthcare}
\label{subsec_literature_hc}
\citet{wolff2019probast} developed a critical appraisal tool called the Prediction Algorithm Risk of Bias Assessment Tool (PROBAST), to assess the risk of bias and applicability of diagnostic and prognostic prediction algorithms. This tool poses a set of questions across four domains to identify biases, with the \textit{participant domain} being particularly relevant to SSB. It considers to what extent the study and the target populations match. A question specifically related to SSB is ``\textit{1.2 Predictors domain: Were all inclusions and exclusions of participants appropriate?}''.
Several reviews and systematic reviews have recently employed PROBAST to study the risk of bias in risk prediction studies, treating causal and machine learning risk prediction studies similarly. For example, 
\citet{Navarron2021} conducted a systematic review of different biases in machine learning prediction techniques. Out of 152 studies, 63 had a study population not representative of the target population or lacked information for verification, and 89 did not handle censoring and related events or had no available information.
\citet{Jong2021appraising} performed a meta-review of 50 reviews, with 2104 models and found that information on the risk of bias per domain was available for 1039 (47\%) studies, of which 25\% had a high risk of bias in participant selection.

To address SSB in machine learning for healthcare, some studies have developed study-specific techniques. For example, \citet{thakur2018use} identified selection bias in early-stage breast cancer prediction due to different observers scoring different regions of the same tumour. They proposed an automated method using whole-slide analysis to mitigate the bias. Similarly, \citet{mei2019knowledge} considered selection bias due to the secondary use of regional electronic health records data and proposed a knowledge learning symbiosis framework to incorporate domain knowledge to address the bias. Additionally, in medical statistics and clinical studies, SSB is commonly treated as a missingness problem under missingness-at-random. Missing outcomes, along with other missing attributes of a patient, are imputed using multiple imputations. For example, \citet{liu2023combining} used imputation for post-menopausal breast cancer in the UK Biobank.

A search on PubMed\footnote{search performed on February 01, 2024.} for ``(risk prediction) AND (machine learning)'' lists 14,023 articles. In contrast, a search for ``(risk prediction) AND (machine learning) AND ((selection bias) OR (participant bias))'' lists only 138 articles, of which 81 were reviews/systematic-reviews/meta-reviews, 45 did not discuss selection bias, seven were irrelevant, two were related to causality and only three discussed SSB with solutions specific to their problem settings.
Recently, \citet{yu2020one} reviewed SSB in radiology and discussed two approaches to address SSB. First to use an independent external test dataset to determine if SSB exists, and second to provide as much detail as possible about the target population and the study population. Interestingly, both approaches are based on preventing SSB. These instances highlight that despite the significance of SSB, machine learning for healthcare has scarcely studied SSB.

\subsection{SSB in other Fields}
\label{subsec_literature_others}
SSB was initially discussed in econometrics and received great attention \cite{berk1983introduction,heckman1990varieties,vella1998estimating} because in econometrics data collection often relies on surveys where participants are self-selected and are not representative of the population \cite{gronau1974wage}. \citet{heckman1979sample}, in a Nobel Prize-winning work, developed a method to correct SSB, using the probability of selection into the sample. However, this method is only applicable to linear regression algorithms. \citet{zadrozny2004learning} formalised SSB in machine learning and proposed a bias correction technique for classifiers. They used selection ratio $P(s=1)/P(s=1|\mathbf{x})$ as a weight for each example in a cost-sensitive learning approach \cite{elkan2001foundations,vogel2020weighted}. This is similar to inverse propensity score or inverse probability of weighting (IPW) as used in treatment effects estimation for reweighting examples \cite{hernan2023causal}. IPW is known to destabilise the training for extreme probabilities. Recently, \citet{Alaimo2023} employed IPW for SSB correction in tax audits and \citet{bradley2022addressing} employed IPW for SSB correction in the UK Biobank dataset for neurological imaging cohort. \citet{huang2006correcting} proposed kernel mean matching (KMM) which does not need selection probabilities, and performs reweighting of the study population such that the means of the study population and the target population are close in a reproducing kernel Hilbert space. Later, \citet{cortes2008sample} presented an SSB correction theory based on distributional stability and analysed bias correction techniques, including KMM.

As discussed in Section~\ref{sec_problem_def}, covariate shift or domain adaptation is closely related to SSB and techniques developed in domain adaptation can be applied to address SSB. Domain adaptation techniques can be broadly categorised as feature-based learning and instance-based learning \cite{de2021adapt}. Feature-based learning approach for domain adaptation aligns representations of the study and target populations in the latent space, e.g., \citet{ganin2016domain} proposed domain-adversarial training of neural networks (DANN) where patient representations are learned agnostic to selection of patient into the study. On the other hand, the instance-based learning approach for domain adaptation reweights loss function to achieve a balance between the study and the target study distributions, e.g., \citet{sugiyama2007direct} proposed Kullback–Leibler importance estimation procedure (KLIEP) that finds weights which minimise the Kullback-Leibler divergence between the study and the target distributions.

In healthcare, SSB is discussed mostly in the context of causality \cite{Breen_2015,smith2020selection,long2022sample,kundu2023framework}, which is a fundamentally different subject to machine learning as machine learning generally applies a correlational approach \cite{pearl2018book}. Causality has some overlaps with machine learning as it employs statistical algorithms for causal effects estimation of interventions but these problems are different than prediction problems \cite{chauhan2024dynamic}. Causality has developed several techniques to address SSB, including causal diagrams \cite{greenland1999causal} and inverse propensity score weighting \cite{weuve2012accounting}. \citet{banack2019investigating} discusses SSB in ageing research due to survival bias and loss to follow-up for geriatrics research in a causal setting and presents an SSB toolkit consisting of different approaches to address SSB in causality, including causal diagrams, instrument variables, multiple imputation and sensitivity analysis.

Fairness in machine learning predictions is another area linked to SSB, as biased data can perpetuate discriminatory outcomes. However, fairness considers biases based on certain sensitive attributes only. For example, \citet{Du2022fair} discusses SSB for fairness when dependent variable values of a set of samples from training data are missing as a result of another hidden process. Their framework adopts the classic Heckman model \cite{heckman1979sample} for bias correction with Lagrange duality, achieving fairness in regression setting based on a variety of fairness notions. Similarly, to address systematic biases in risk prediction across demographic groups, \citet{Hong2024} employed imbalance learning, transfer learning, and federated learning techniques.

From the above brief discussion, we find that SSB is well recognised as a threat to the generalisation of machine learning algorithms. Despite one of the major hurdles in the clinical adoption of machine learning algorithms, SSB is scarcely discussed. Moreover, the existing techniques to address SSB are based on bias correction which forces alignment of the distributions between the study and the target populations and may lead to a loss of predictive performance of the prediction algorithms. Thus, there is a need to highlight the potential dangers of SSB in machine learning for healthcare to encourage further research and develop techniques that effectively leverage available data to provide accurate predictions, despite the presence of bias.

\section{Methods}
\label{sec_methods}
In this section, we present the proposed target population identification (TPI) approach -- a new research direction -- to address SSB in machine learning for healthcare for developing algorithms that utilise the available biased data to make accurate predictions. The key idea of our proposed approach is that we constrain the underlying machine learning algorithms, at deployment time, to a subpopulation seen by the algorithm during the training process, i.e., to a target subpopulation which is representative of the study population, without requiring to balance distributions of the two populations. The non-selected patients are referred to a clinician, ensuring they receive appropriate care while maintaining the reliability of algorithmic predictions. This differs from the existing SSB correction approaches which forces alignment of the distributions of the study and the target populations, resulting in a loss of predictive performance of algorithms.
Our novel approach is motivated by the observation that in healthcare the study and the target populations differ in terms of the non-selected patients. This is unlike domain adaptation where the target dataset is different to the source dataset and the aim is to perform better on the target dataset using information from the source dataset \cite{kouw2019review}. So for healthcare, an approach based on the identification of the target subpopulation representative of the study population can help in addressing SSB.

\begin{figure}[htb!]
    \centering
    \includegraphics[width=0.6\linewidth]{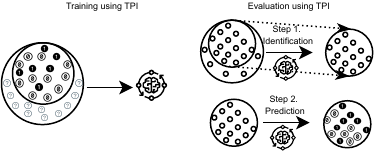}
    \caption{An illustration of the proposed target population identification (TPI) approach which employs a two-step evaluation process where first it identifies the target subpopulation representative of the study population, followed by predictions for the selected subpopulation. The non-selected patients are referred to a clinician.}
    \label{fig_TIA}
\end{figure}

At training time (or development time), the proposed TPI approach, in addition to learning the underlying risk prediction task from the study population, also learns the selection process of including patients in the study population using patient attributes, i.e., $P(s=1|\mathbf{x})$ using included as well as excluded patients. At test time (or deployment time), the proposed approach employs a two-step evaluation process. First, it identifies the target patient subpopulation that is representative of the study population, and then it applies the risk prediction algorithm to the patients identified in the first step, as depicted in Fig.~\ref{fig_TIA}. Thus, the proposed approach is based on the assumption that the selection process can be identified from patient characteristics $\mathbf{x}$. In other words, the proposed approach assumes that the outcomes of the non-selected patients are missing-at-random and it refers to case (ii) as discussed in Section~\ref{sec_problem_def}. Moreover, the learning of the selection process has access to more data than the underlying risk prediction task because it considers included as well as excluded patients.

To implement the TPI approach, we propose two techniques called T-Net and MT-Net, as described in the following subsections.

\subsection{T-Net}
\label{subsec_TNet}
\begin{figure}[htb!]
    \centering
    \includegraphics[width=\linewidth]{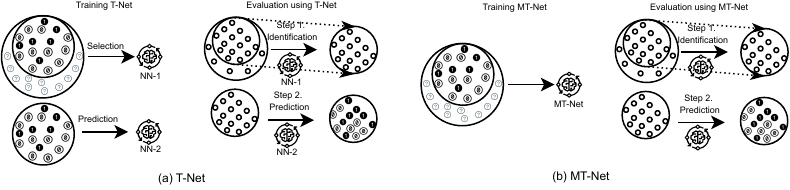}
    \caption{An illustration of the proposed T-Net and MT-Net, as specific techniques under target population identification (TPI), to address SSB. T-Net trains two independent neural networks (NN) while MT-Net uses a single multitasking network.}
    \label{fig_nets}
\end{figure}

To address SSB using the TPI approach, T-Net develops two independent neural networks, one for identification, i.e., to predict the selection/inclusion of a patient in the study population and the other network for the risk prediction task. The two networks can be trained in any order. The selection network trains a classifier $s=f:\mathbf{x} \rightarrow \{0,1\}$ over patients included as well as excluded from the study, where $s=1$ for patients included in the study population, i.e., $\mathbf{x} \in \mathbf{X}^1$ and $s=0$ for patients excluded from the study due to missing outcomes, i.e., $\mathbf{x} \in \mathbf{X}^0$. On the other hand, the risk prediction network trains a separate neural network $y=g:\mathbf{x} \rightarrow y$ over the study population (only patients included in the study), i.e., $\mathbf{x} \in \mathbf{X}^1$.
T-Net employs a two-step evaluation where first the selection network identifies the target subpopulation belonging to the study population, followed by predictions by the risk predictor for the patients selected in the first step. Fig.~\ref{fig_nets}(a) depicts the working of T-Net.
Conversely, employing a naive machine learning algorithm would involve training the risk predictor $g(x)$ solely on data from selected patients, intending to make predictions for the entire target population. However, this approach may result in misclassifications for patients within the non-selected subpopulation, potentially leading to unnecessary or inadequate interventions. The risk predictor, developed on a biased dataset, lacks information about the excluded subpopulation, contributing to these shortcomings.

Hence, T-Net makes predictions specifically for the subpopulation on which it was trained, leveraging available information. In contrast, a naive algorithm is compelled to make predictions for a subpopulation it may not have encountered during training, leading to potential inaccuracies due to lack of relevant information.

\subsection{MT-Net}
\label{subsec_MTNet}
To address SSB using the TPI approach, MT-Net develops a multitasking neural network which has representation layers $\phi: \mathbf{x} \rightarrow \mathbb{R}^{d'}$, learning shared patient representation of $d'$ dimensions, followed by two task-specific heads, one to predict the selection $s=f:\phi(\mathbf{x}) \rightarrow \{0,1\}$ and the other for risk prediction $y=g:\phi(\mathbf{x}) \rightarrow y$. Thus, MT-Net facilitates shared learning between the selection and the risk prediction and benefits from inductive learning (knowledge transfer between tasks with a shared domain) between the tasks.
Similar to T-Net, MT-Net employs a two-step evaluation process where the selection task identifies the subpopulation of the target population which is representative of the study population, followed by predictions by the risk prediction task for the patients selected in the first step. Fig.~\ref{fig_nets}(b) depicts the working of MT-Net.
On the other hand, a naive multitasking approach, which employs one task each for selection and risk prediction, benefits from shared learning but predicts for the entire target population. Similar to the naive single-task risk predictor, a naive multitasking predictor might lead to unnecessary or insufficient interventions because it's trained on the biased study population.
Similar to T-Net, MT-Net makes predictions specifically for the subpopulation on which it was trained. In contrast, a naive multitasking algorithm is compelled to make predictions for a subpopulation it may not have encountered during training, leading to potential inaccuracies due to a lack of relevant information.

Therefore, both TPI approaches, T-Net and MT-Net, train on both the selected and the non-selected patients, and employ a two-step evaluation. This ensures predictions are made for the target subpopulation encountered during training, yielding similar performances on both study and target populations -- essential for the development and deployment of machine learning algorithms. Thus, the proposed approach not only offers a novel method for addressing SSB but also utilises the data which otherwise is not selected and is particularly crucial for data-limited studies. While T-Net provides flexibility with two separate networks, MT-Net benefits from shared learning, proving advantages in settings with limited data for tackling SSB. The architectures of T-Net and MT-Net follow standard single-task and multitasking networks in deep learning, respectively. The number of layers and neurons in each of the layers are described in Section~\ref{subsec_exp_settings} under the heading `Hyperparameters and Performance Metrics'.

\section{Results}
\label{sec_results}
This section discusses the experimental setup of the empirical study and presents results with synthetic and semi-synthetic datasets to analyse the performance of different techniques to address SSB.

\subsection{Experimental Setting}
\label{subsec_exp_settings}
\textbf{Baselines:} We have utilised commonly used baselines, including IPW and imputation, and baselines from machine learning, including SSB handling and domain adaptation techniques, as given below.\\
\textit{Oracle:} It presents a case where the study and the target population have the same distributions, and its performance reflects the ideal situation without any bias. Here, an algorithm is trained on a biased dataset and tested on a biased dataset, i.e., the algorithm has seen the target population during the training time. So, the closer the performance of the rest of the algorithms to Oracle the better it is.\\
\textit{Naive:} As discussed in Section~\ref{sec_methods}, a naive approach ignores SSB and develops an algorithm with a biased dataset and evaluates with an unbiased dataset as the algorithm does not have any mechanism to correct the bias or identify the target population. So, Naive should exhibit the lowest performance in the presence of SSB. Naive, Oracle and T-Net methods have same shared neural network for risk prediction, however, the three differ in their evaluation process. Naive is evaluated on the entire test dataset, Oracle is evaluated only for patients belonging to the study population, and T-Net identifies the target subpopulation representing the study population and is evaluated only for the identified patients.\\
\textit{MT-Naive:} It is a naive multitasking technique to address SSB which similar to MT-Net has two tasks for the selection and the risk prediction. Similar to Naive, it trains the risk prediction task on the biased dataset and is evaluated on the unbiased dataset.\\
\textit{IPW:} It first learns the selection probability for each data point followed by cost-weighted training for risk prediction where each data point is weighted by the inverse of the selection probability, as discussed above Section~\ref{sec_related_work}.\\
\textit{Imputation:} This is the most commonly used approach in statistical and clinical literature to address SSB. This treats SSB as a missingness handling problem to impute outcomes for the non-selected patients, followed by a standard supervised machine learning approach. We utilise the state-of-the-art MissForest algorithm \cite{Daniel2012MissForest} for imputation which is based on multiple imputations and utilises random forest.\\
\textit{KMM:} It performs reweighting of the study populations such that the means of the study populations and the target populations are close in a reproducing kernel Hilbert space.\\
\textit{KLIEP:} It also performs reweighting but finds weights which minimise the Kullback-Leibler divergence of the study and the target study distributions.\\
\textit{DANN:} It learns patient representations agnostic to the selection of patients into the study through adversarial training. This technique, similar to MT-Net, has a multitasking network and access to the included as well as excluded patients, however, the selection network is used for adversarial training and not for the identification.

\textbf{Datasets:} For \textit{synthetic data generation}, following the literature \cite{chauhan2023adversarial}, we generated a dataset with 25 features, slightly more than in real-world datasets discussed below, where features $\mathbf{X}$ are sampled from a uniform distribution over (-10, 10). For a uniformly sampled vector $\mathbf{a}$ of dimensions equal to features and a scalar $c$, outcomes $\mathbf{y}$ for risk prediction are generated using the following process,
\begin{equation}
    \label{eq_DGP_y}
    \mathbf{y} = \left(\mathbf{X}\mathbf{a} + c\right) > 0.
\end{equation}
To add noise to outcomes, we randomly flip one percent of the outcomes to reflect real-world data quality issues and imperfections. For introducing SSB to the dataset, we take another uniformly sampled vector $\mathbf{b}$ and a scalar $d$, and generate selection variable $\mathbf{s}$ using the following process,
\begin{equation}
    \label{eq_DGP_s}
    \begin{array}{l}
    \mathbf{s} = \mathbf{y} \oplus \left(\left(\mathbf{X}\mathbf{b} + d\right) > 0\right),
    \end{array}
\end{equation}
where $\oplus$ represents an exclusive OR operation. For \textit{semi-synthetic datasets}, we have access to real-world features as well as outcomes. To generate a selection variable, we follow \cite{cortes2008sample}. For a random projection vector $\mathbf{a} \in[-1,1]^d$ of dimensions equal to the number of features, and mean of features $\mathbf{\bar{x}}$, the selection probability of a patient $\mathbf{x}$ is calculated as given below.
\begin{equation}
    \label{eq_DGP_bias_gen}
    \begin{array}{ll}
    p(s=1|\mathbf{x}) =& \dfrac{1}{1 + e^{-\mathbf{v}}}, \;\;\;\;
    \text{where } \mathbf{v} = \dfrac{4 \mathbf{a}.\left(\mathbf{x}-\bar{\mathbf{x}}\right)}{\sigma_{\mathbf{a}.\left(\mathbf{x}-\bar{\mathbf{x}}\right)}},
    \end{array}
\end{equation}
where $\sigma_{\mathbf{a}.\left(\mathbf{x}-\bar{\mathbf{x}}\right)}$ is standard deviation of expression $\mathbf{a}.\left(\mathbf{x}-\bar{\mathbf{x}}\right)$.

We used two semi-synthetic datasets, hereon called Diabetes and COVID-19 datasets. Diabetes dataset is a health-related telephone survey dataset\footnote{https://www.kaggle.com/datasets/cdc/behavioral-risk-factor-surveillance-system}, collected annually by the Behavioral Risk Factor Surveillance System at the Centers for Disease Control and Prevention. We have used a cleaned, consolidated and balanced version from Kaggle\footnote{https://www.kaggle.com/datasets/alexteboul/diabetes-health-indicators-dataset}, and the dataset is a binary dataset to predict diabetes, and it has 70,692 data points, 21 features with an event rate of 50\%. COVID-19 dataset is an anonymised dataset released by the Mexican government\footnote{https://www.gob.mx/salud/documentos/datos-abiertos-152127} and accessed from Kaggle\footnote{https://www.kaggle.com/datasets/tanmoyx/covid19-patient-precondition-dataset/data}. Since the dataset has high missingness, we have excluded records with missing values. This resulted in a complete dataset comprising 20,352 data points and 18 features, used to predict whether an individual has COVID-19 or not, with an event rate of 41\%.

\textbf{Hyperparameters and Performance Metrics:}
We employed neural networks for analysing SSB due to their flexibility, expressiveness, and popularity \cite{chauhan2023brief,bishop2024DL}, and the fact that SSB handling methods are mostly implemented with neural networks \cite{huang2006correcting,ganin2016domain} and there is no direct way to implement multitasking with standard machine learning algorithms. We split the datasets as 20\% of the full dataset as a test dataset, 20\% of the remaining as a validation dataset and the rest as a training dataset. We utilised Adam optimiser \cite{kingma2014adam} and tuned the learning rate from 0.0001 and 0.0005. A mini-batch size of 64 data points and 1000 epochs were used, and an early stopping with patience of 10 epochs. A dropout rate of 10\% served as regularisation to prevent overfitting. For all the methods, the neural network architectures were tuned for hidden layers from [50], [100], [100, 100], where the number of list elements indicates the number of hidden layers, and values represent the number of neurons in the corresponding hidden layer. For task-specific layers, we tuned from [50], [100].
For a performance metric, we use the Area Under the Receiver Operating Characteristic Curve (AUC) to assess an algorithm's ability to distinguish between positive and negative instances. It quantifies the area under the curve generated by plotting the true positive rate against the false positive rate at various classification thresholds. A higher AUC (closer to 1) indicates better discrimination performance, while a random classifier has an AUC of 0.5.

All the experiments are implemented in Python using PyTorch \cite{paszke2019pytorch}. The experiments are executed on an Ubuntu machine (64GB RAM, one NVIDIA GeForce GTX 1080) and each experiment is averaged over ten seeds.

\subsection{Analysis with the Synthetic Data}
\label{subsec_synthetic}
\begin{figure}[htb!]
    \centering
    \includegraphics[width=0.65\linewidth]{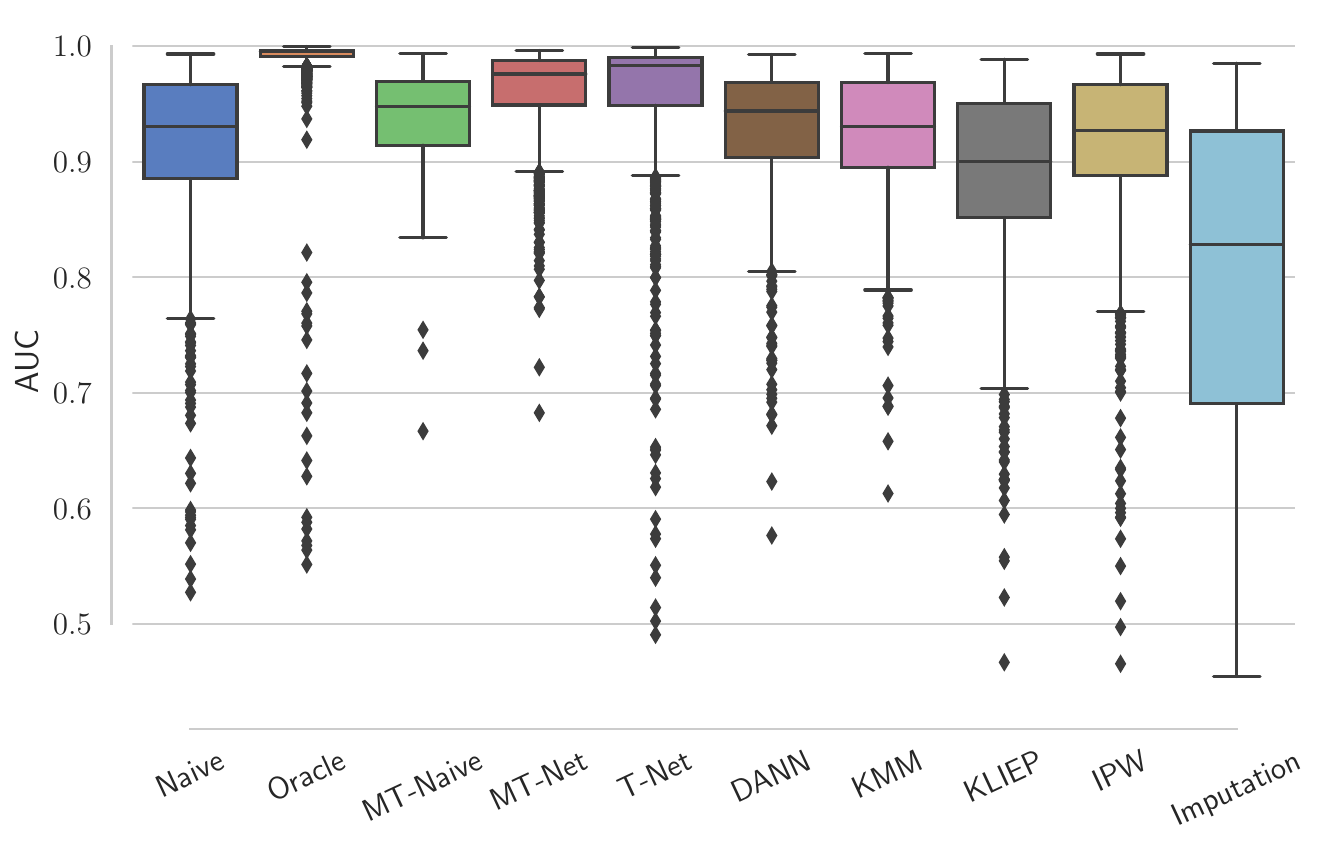}
    \caption{Comparative analysis of the average performance of various techniques for handling SSB using the synthetic dataset across the event rates and the non-selection rates ranging from 5\% to 40\%, and dataset sizes ranging from 1000 to 5000.}
    \label{fig_synthetic_summary}
\end{figure}

Here, we present a comparative study of various techniques for handling SSB with the synthetic dataset. Since the performance could vary across different event rates, selection rates or dataset sizes, first we present a summary of the performance across different settings using boxplots in Fig.~\ref{fig_synthetic_summary}, and subsequent subsections study the effect of the selection/event rates and dataset sizes. We consider the event rates and the non-selection rates from [5, 10, 20, 30, 40]\% as we don't expect large non-selection rates or event rates in real-world datasets, and dataset sizes from [1000, 2000, 3000, 4000, 5000]\footnote{For a dataset size of 1000 with a non-selection rate and an event rate of 10\% each, it implies there are 100 non-selected patients and 900 selected patients. 90 Of the 900 selected patients have positive outcomes while the rest 810 has negative outcomes.}, where each setting is repeated and averaged over 10 seeds. We observe large variations in the performance of all the techniques due to diverse settings, where extreme cases like low event rates, low non-selection rates or small dataset sizes could lead to a large drop in the performance.

Upon analysing the performance of various techniques, we note an average drop of over seven percent for Naive technique compared to Oracle, which is indicative of reduced performance on the target population compared to the study population. This is because Oracle is evaluated on the target population which is representative of the study population, however, Naive has SSB.
Oracle performs the best in terms of median performance, as it represents a bias-free setting, and Imputation technique performs the worst in terms of median performance as well as variation in the performance.
Oracle is followed by the proposed T-Net and MT-Net, however, MT-Net shows the least variation in the performance among all the techniques, including Oracle, across all the settings. The smaller variation in the performance of MT-Net occurs because it has access to included as well as excluded patients and it employs shared learning between the tasks, which is helpful with limited data settings.
Interestingly, the performance of KLIEP and Imputation fall below Naive technique which indicates that the bias handling negatively impacts the performance. Also, a naive multitasking, i.e., MT-Naive improves over Naive and performs better than the rest of the baselines. We study the effect of the non-selection rates, the event rates and dataset sizes on the performance of various SSB handling techniques in the following subsections.

\subsubsection{Effect of the Non-selection Rates and the Event Rates}
\label{subsubsec_syn_effect_cr}
\begin{figure}[htb!]
     \centering
     \begin{subfigure}[b]{0.32\textwidth}
         \centering
         \includegraphics[width=\textwidth]{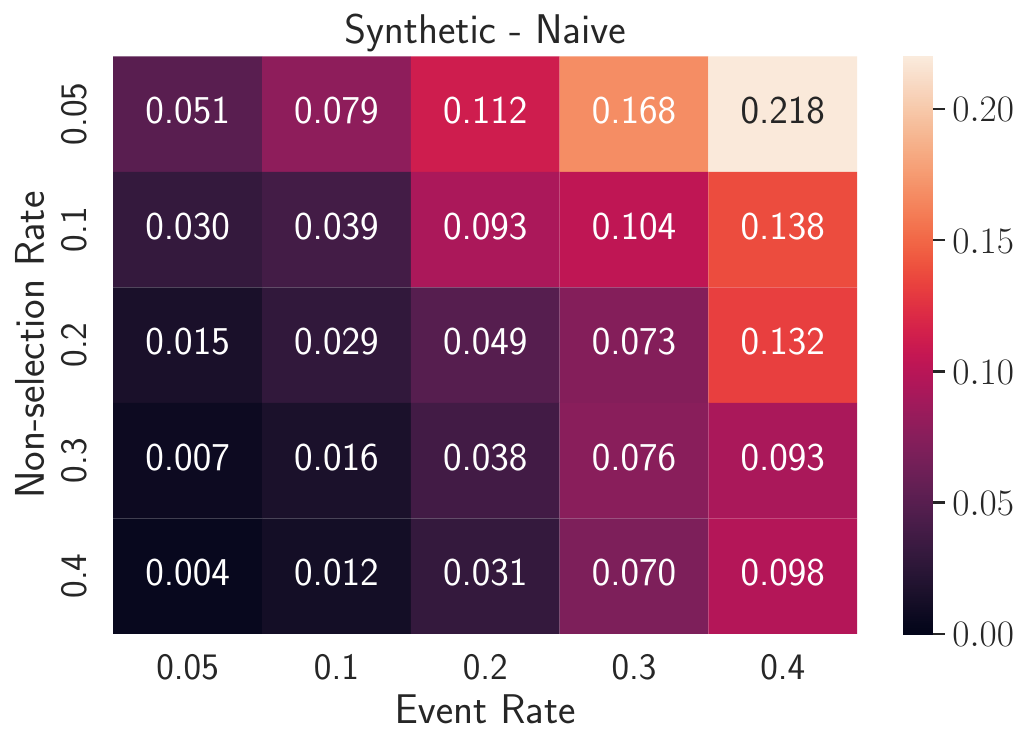}
     \end{subfigure}
    ~
    \begin{subfigure}[b]{0.32\textwidth}
         \centering
         \includegraphics[width=\textwidth]{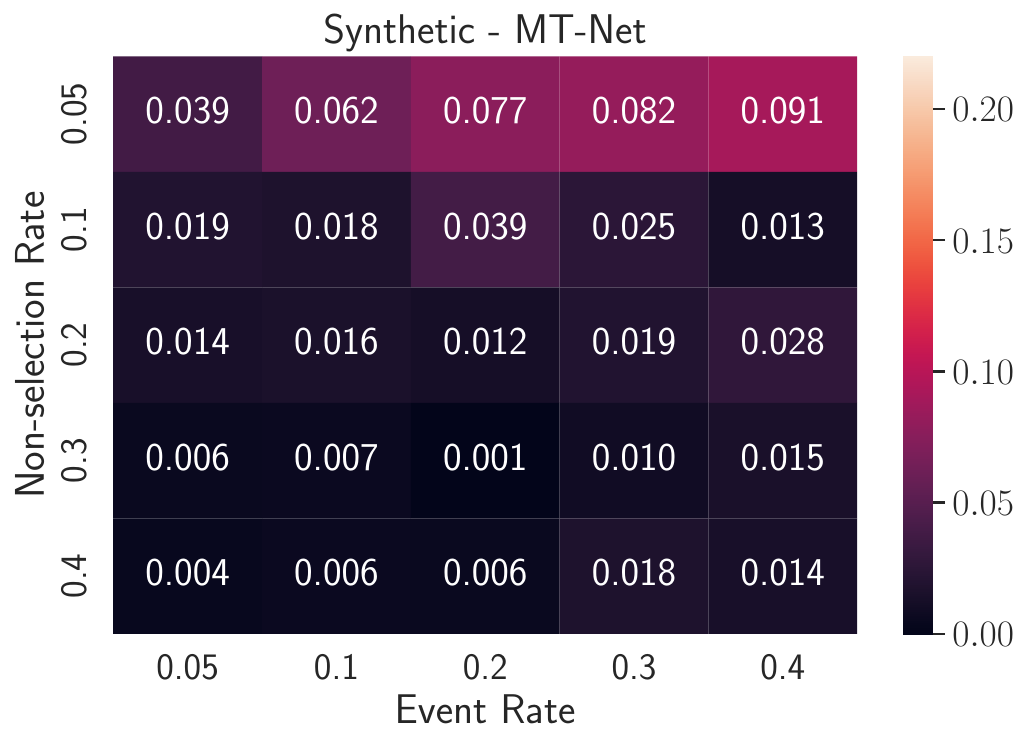}
     \end{subfigure}
    ~
     \begin{subfigure}[b]{0.32\textwidth}
         \centering
         \includegraphics[width=\textwidth]{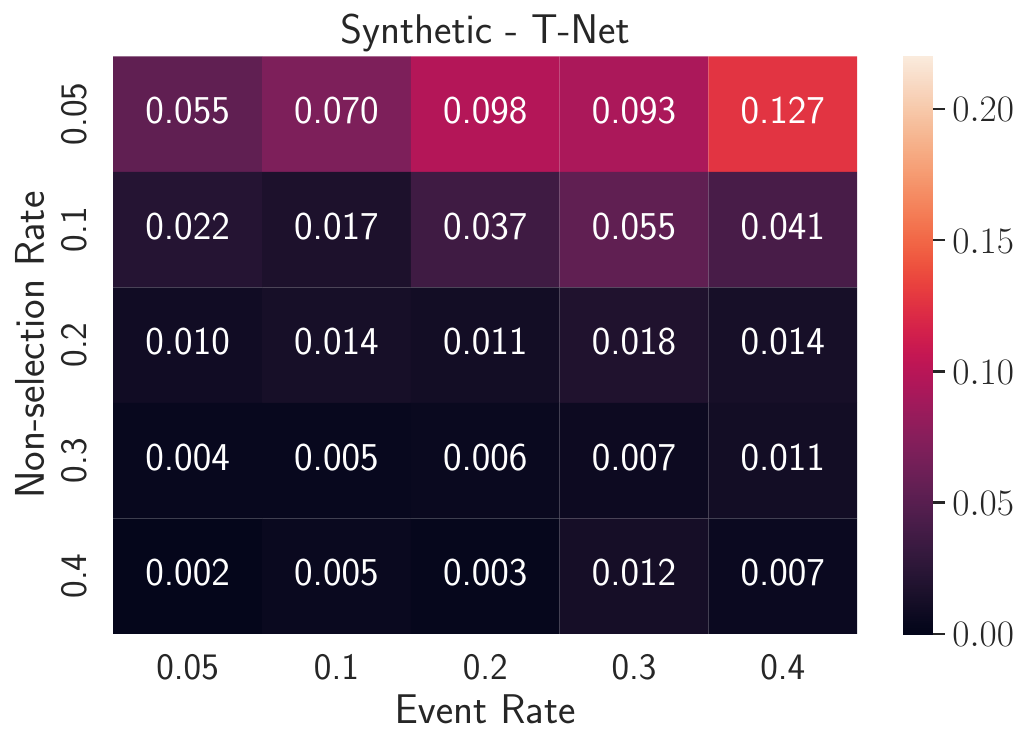}
     \end{subfigure}
     
     \begin{subfigure}[b]{0.32\textwidth}
         \centering
         \includegraphics[width=\textwidth]{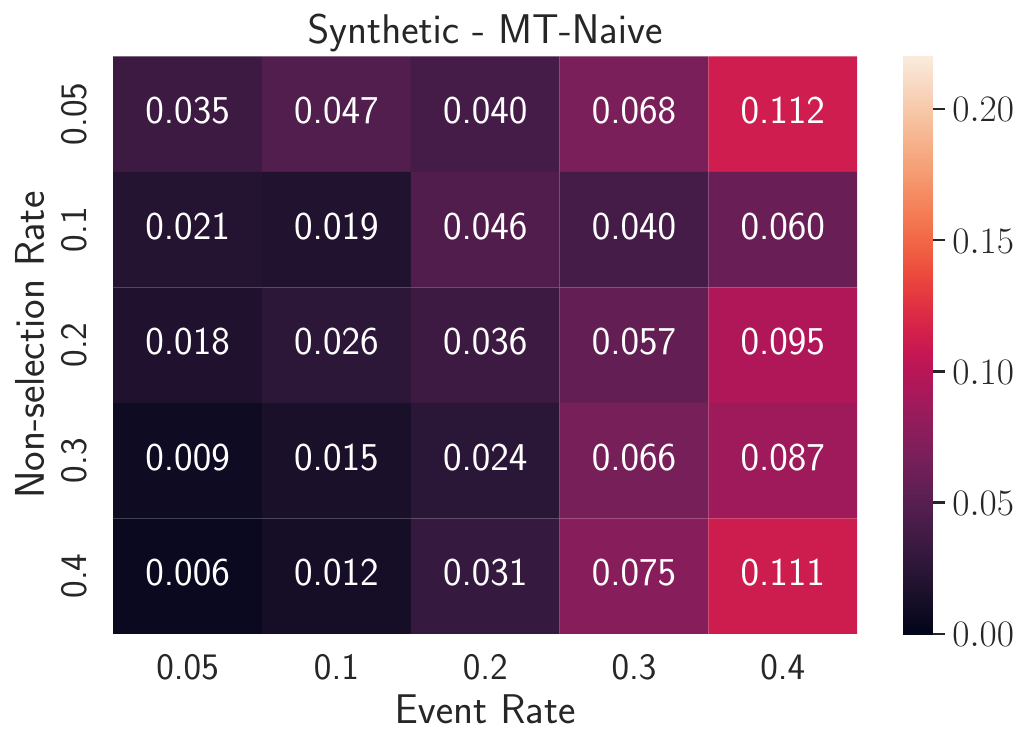}
     \end{subfigure}
     ~
     \begin{subfigure}[b]{0.32\textwidth}
         \centering
         \includegraphics[width=\textwidth]{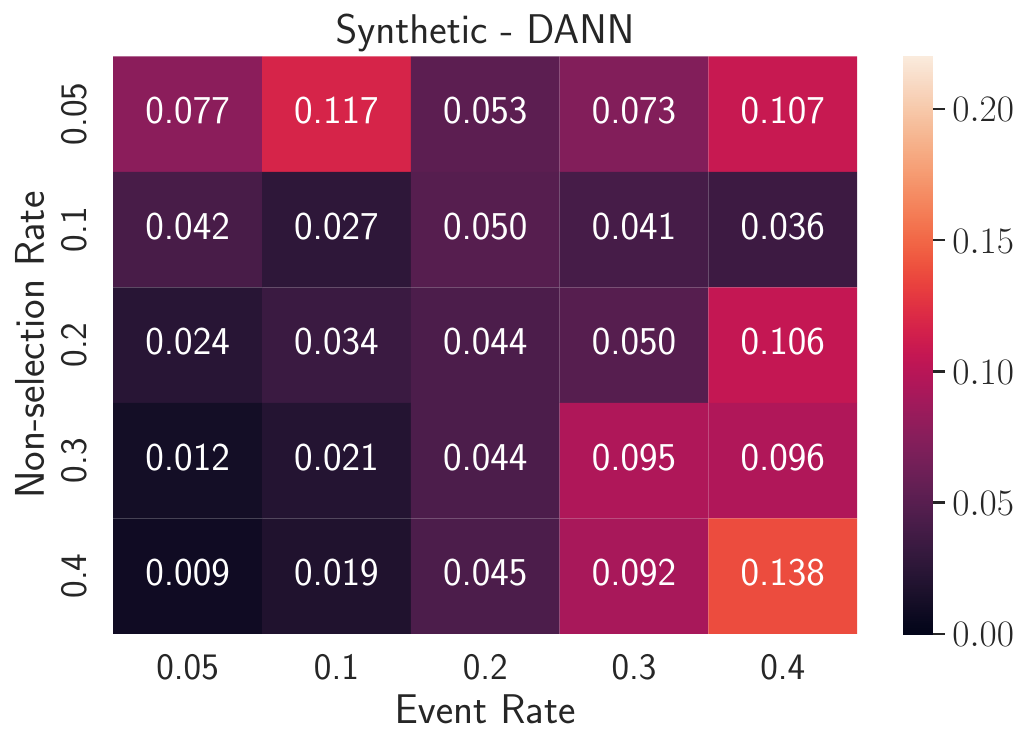}
     \end{subfigure}
    ~
     \begin{subfigure}[b]{0.32\textwidth}
         \centering
         \includegraphics[width=\textwidth]{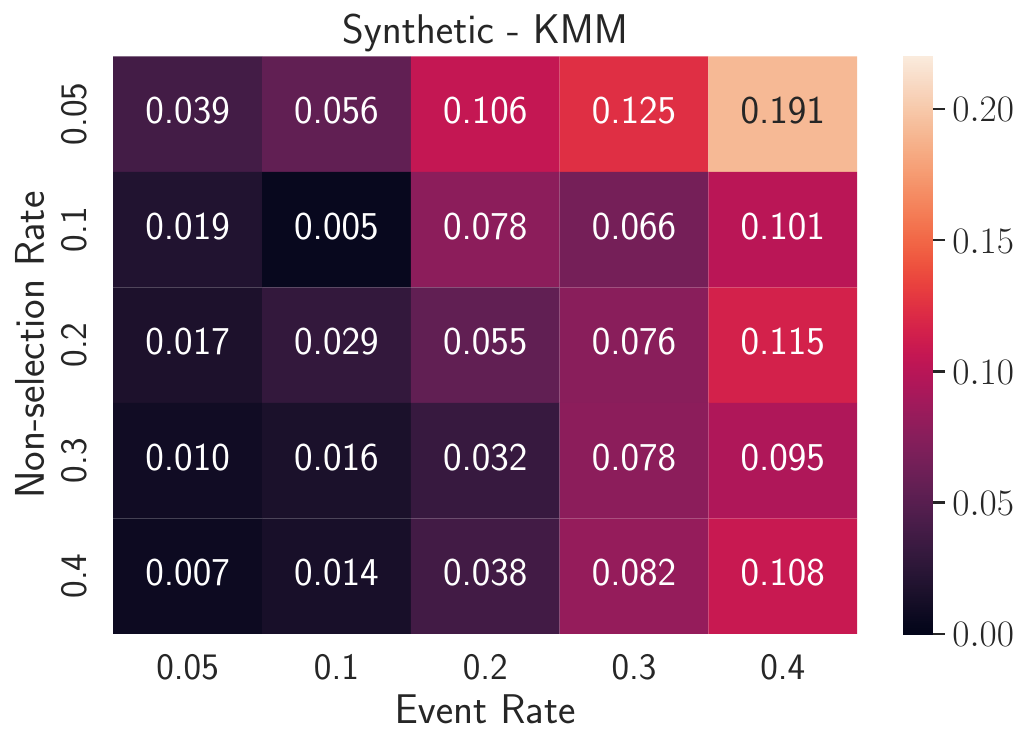}
     \end{subfigure}
     \caption{Effect of the event rates and the non-selection rates on the performance of different techniques to address SSB using the synthetic dataset, as averaged over dataset sizes ranging from 1000 to 5000.}
     \label{fig_synthetic_risk_vs_selection}
\end{figure}

To study the effect of the non-selection rates and the event rates on the performance of different techniques to address SSB, we consider the event rates and the non-selection rates from [5, 10, 20, 30, 40]\%, and present a summary of the performance averaged over dataset sizes from [1000, 2000, 3000, 4000, 5000]. For the sake of clarity, we consider the proposed techniques and two best baselines, i.e., DANN and KMM, along with the naive techniques, and present absolute differences in their performance from Oracle as heatmaps in Fig.~\ref{fig_synthetic_risk_vs_selection} because Oracle presents an ideal case and unbiased model.

From Fig.~\ref{fig_synthetic_risk_vs_selection}, it is clear that overall MT-Net and T-Net perform better than the baselines as the difference with Oracle is lesser than the baselines.
The relative performance of Naive technique drops up to 22\% for a low non-selection rate and a high event rate, while the lowest drop occurs at a high non-selection rate and a low event rate. 
This underlies the importance of addressing SSB in machine learning for healthcare as machine learning algorithms showing high performance may not show similar performance when deployed due to SSB.
It is interesting to note that for some settings, like a high non-selection rate and a low event rate, all four techniques perform closely to Oracle and then the relative performance decreases as the non-selection rate decreases or the event rate increases. The baselines show more drop in the performance as compared with the proposed techniques. The worst performance occurs at a low non-selection rate and a high event rate for all techniques, except DANN which shows a slightly different pattern at an event rate of 0.3 and 04. Since we keep the total number of data points fixed while we change one of the two rates, the pattern of a decrease in the relative performance with a decrease in the non-selection rate can be accounted for less available data to address SSB as well as an increase in imbalance for non-selection rate. On the other hand, the pattern of a decrease in the relative performance with an increase in the event rate, i.e., a decrease in class imbalance, for the event rate across all the techniques is a bit counter-intuitive, as generally, the performance improves with improvement in class imbalance. Further analysing, we observe that the performance of individual techniques does increase with an increase in the event rate, however, the relative performance presented in the heatmaps shows a decrease because Oracle shows less variability with event rate than the rest of the techniques resulting in a decrease in the performance with an increase in event rate. These results also show a consistency of the proposed techniques across different settings as compared with the baselines. Overall, Naive performs the worst and MT-Net performs the best to address SSB.

\subsubsection{Effect of Dataset Size}
\label{subsubsec_syn_effect_scale}
\begin{figure}[htb!]
    \centering
    \includegraphics[width=0.65\linewidth]{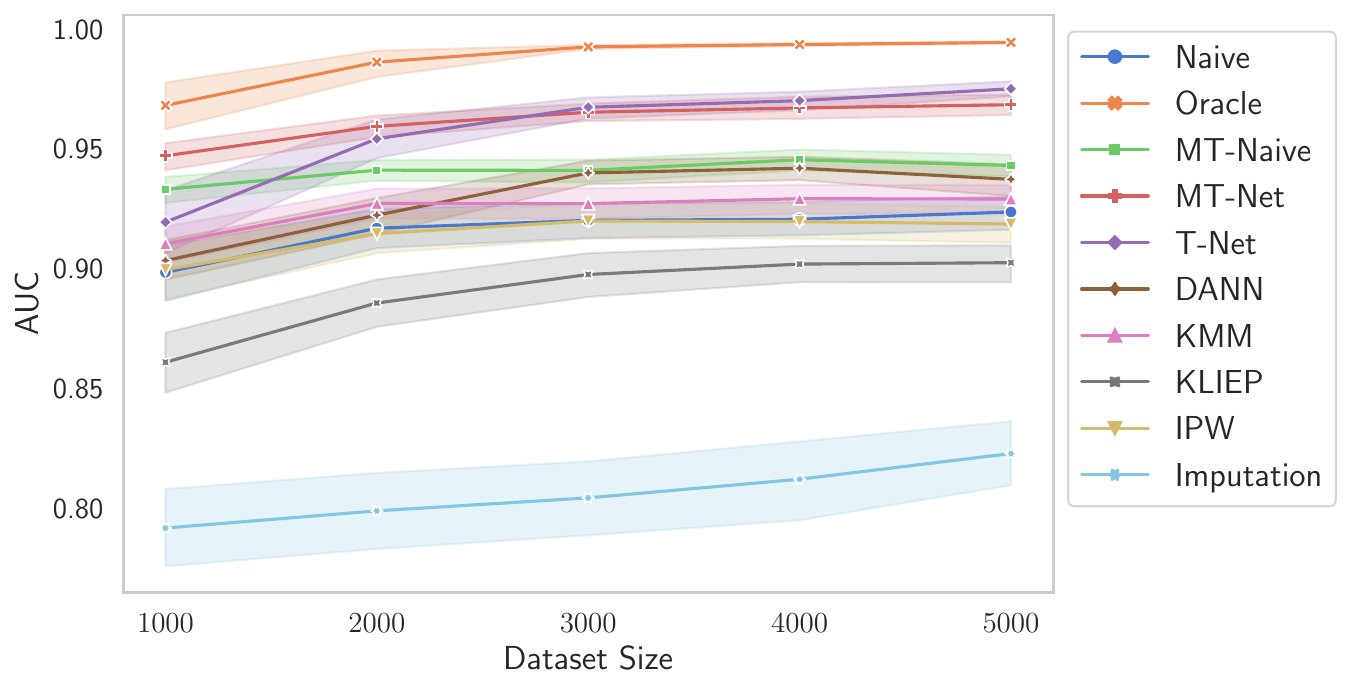}
    \caption{Effect of dataset size on the performance of different techniques to address SSB using the synthetic dataset, as averaged across event rates and non-selection rates ranging from 5\% to 40\% (shaded region represents one standard deviation over 10 runs).}
    \label{fig_synthetic_scale}
\end{figure}

To study the effect of dataset size on the performance of different techniques to address SSB with the synthetic dataset, we consider dataset sizes from [1000, 2000, 3000, 4000, 5000], and present a summary of the performance in Fig.~\ref{fig_synthetic_scale} averaged over the event rates and the non-selection rates from [5, 10, 20, 30, 40]\%.

From Fig.~\ref{fig_synthetic_scale}, we observe that Oracle is the best-performing technique to address SSB, as it is bias-free, while Imputation is the worst-performing technique. MT-Net and T-Net outperform the rest of the baselines across all dataset sizes, except for small datasets where MT-Naive performs better than T-Net, though MT-Net still outperforms MT-Naive. As observed earlier, MT-Net is more effective than T-Net for smaller datasets due to access to the included and the excluded patients and sharing of learning between the tasks. Interestingly, KLIEP and Imputation techniques perform worse than Naive technique which indicates that the bias correction fails to correct the bias and acts negatively resulting in a loss of performance. Another notable observation is that the naive multitasking, i.e., MT-Naive performs better than the rest of the baselines, including Naive technique. Overall, as expected, the performance of all the techniques for addressing SSB improves with an increase in the dataset size, however, once there are sufficient patients, there isn't much improvement. Similarly, deviation in the performance also improves with an increase in dataset size.

\subsection{Analysis with the Semi-synthetic Data}
\label{subsec_real}
\begin{figure}[htb!]
     \centering
     \begin{subfigure}[htb!]{0.49\textwidth}
         \centering
         \includegraphics[width=\textwidth]{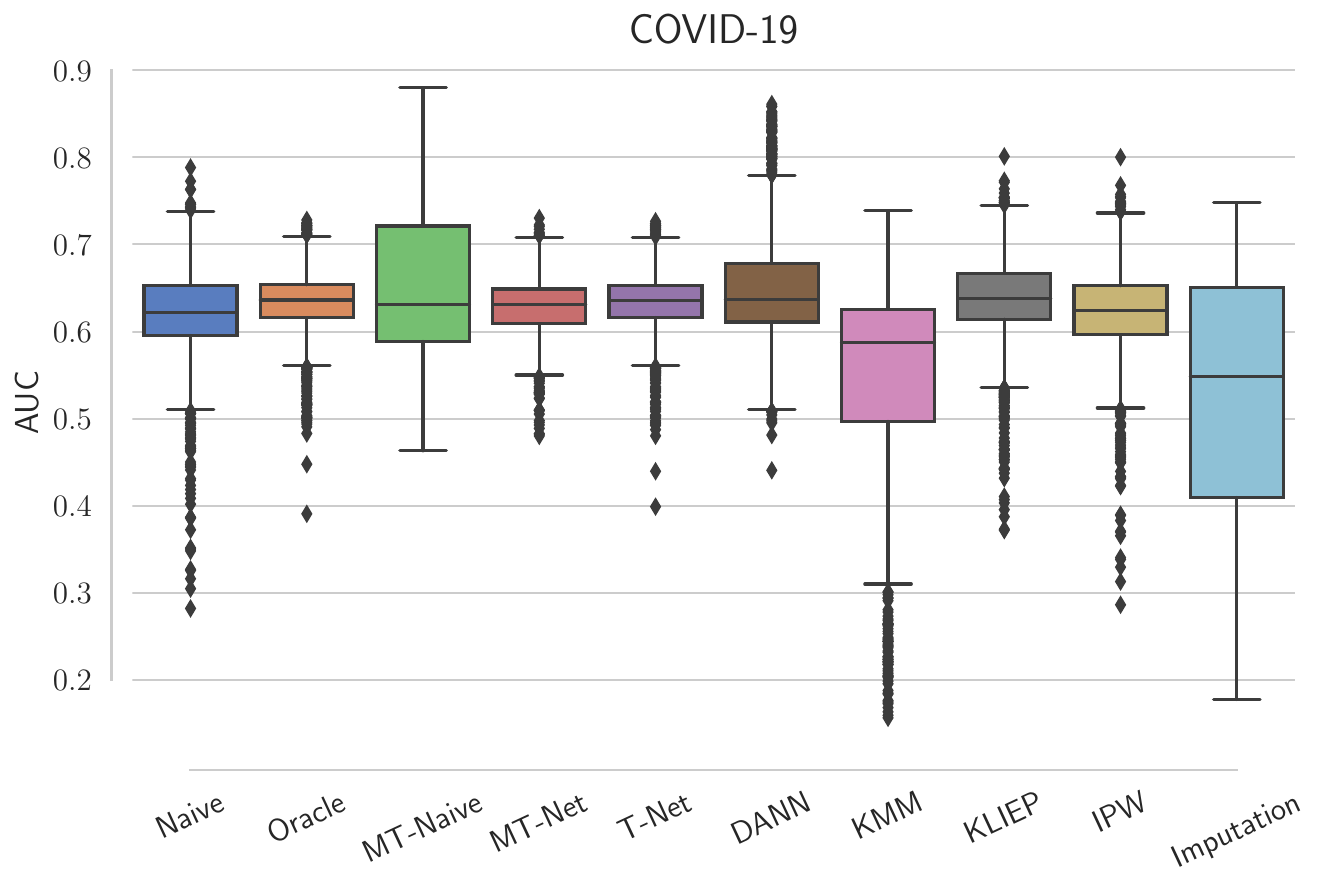}
     \end{subfigure}
    ~
     \begin{subfigure}[htb!]{0.49\textwidth}
         \centering
         \includegraphics[width=\textwidth]{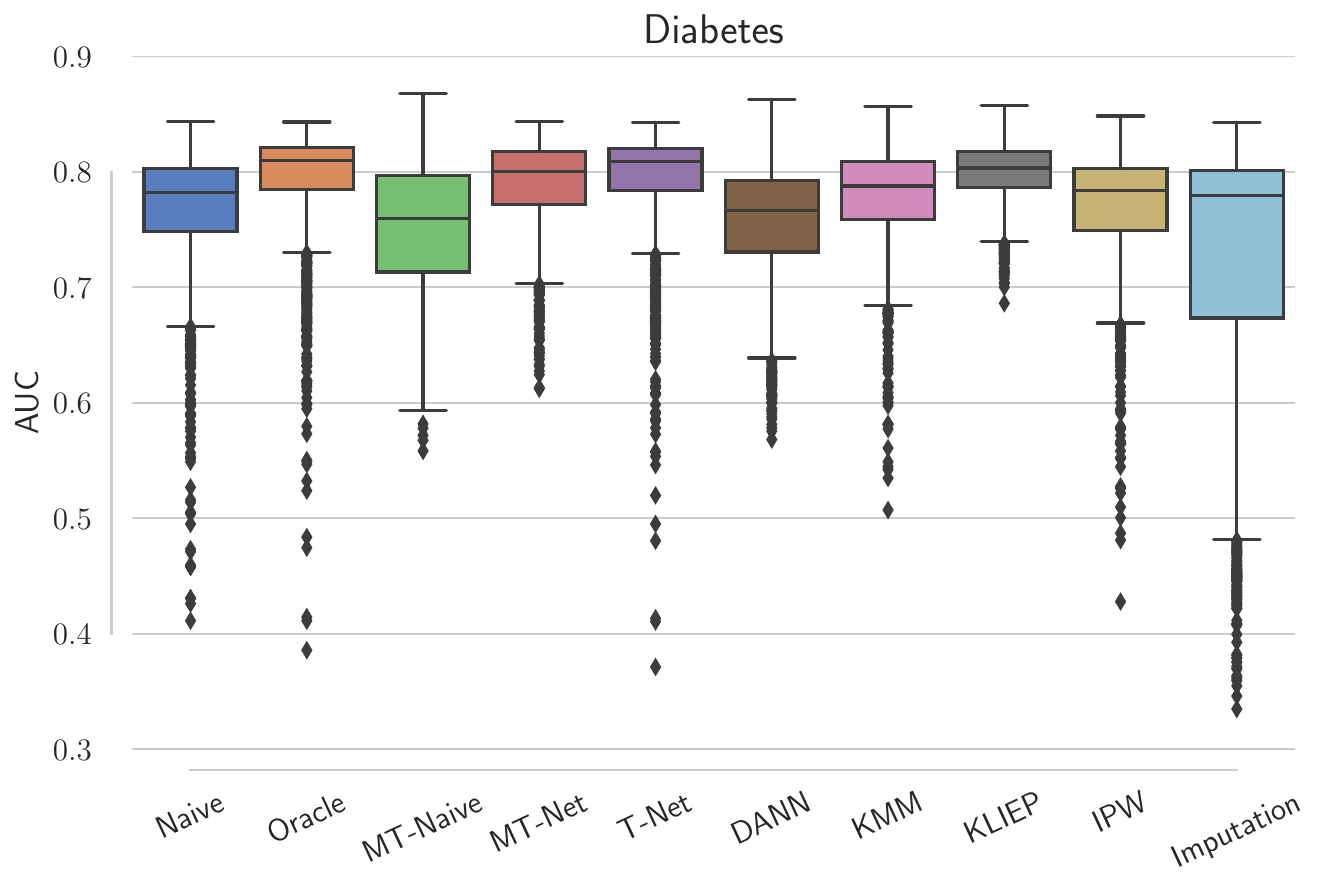}
     \end{subfigure}
     \caption{Comparative analysis of the average performance of various techniques for handling SSB using COVID-19 and Diabetes datasets across event rates and non-selection rates ranging from 5\% to 40\%, and dataset sizes ranging from 1000 to 15,000/25,000.}
     \label{fig_real_summary}
\end{figure}

Here, we present a comparative study of various techniques for handling SSB with two semi-synthetic datasets for COVID-19 and Diabetes. We need semi-synthetic datasets to evaluate the performance because in practice it is not possible to know the outcomes of the non-selected patients, as that's why they were dropped. For the semi-synthetic datasets, features and outcomes come from a real dataset but SSB is artificially introduced as discussed in Section~\ref{subsec_exp_settings}. Similar to the synthetic dataset, first we present a summary of the performance across different settings using boxplots in Fig.~\ref{fig_real_summary}. We consider the event rates and the non-selection rates from [5, 10, 20, 30, 40]\%, and dataset sizes from [1000, 2000, 5000, 10,000, 15,000/25,000] for COVID-19/Diabetes datasets, where each setting is repeated and averaged over 10 seeds. We observe large variations in the performance of all the techniques due to diverse settings, where extreme cases like low event rates, low non-selection rates or small dataset sizes could lead to a large drop in the performance.

On analysing the performance of Naive technique as compared with Oracle, we observe an average drop of two percent and three percent for COVID-19 and Diabetes, respectively. 
Moreover, while Oracle performs the best in terms of median performance for the COVID-19 and Diabetes datasets, we observe some baselines performing close to Oracle. For example, in the Diabetes dataset, KLIEP and DANN show performance like Oracle. The small drop in average performance across different settings and the good performance of some baselines can be attributed to combination of two factors. First, except for Oracle and the proposed techniques, the remaining techniques make predictions for all patients (i.e., included and excluded). Second, for certain combinations of non-selection rates and event rates, the non-selected patients do not influence the decision function of the prediction model (This happens when the non-selected patients lie on the correct side of the decision boundary and away from the decision boundary that they do not have any role in deciding the decision boundary.). Consequently, all techniques that make predictions for all patients consistently make correct predictions for the non-selected patients, which improves their overall performance for all patients. This is also reflected in Fig.~\ref{fig_real_summary}, where the maximum performance of some baselines is even higher than that of Oracle.

MT-Net shows the least variation in the performance among all the techniques across all the settings. The smaller variation in the performance of MT-Net occurs because it has access to included as well as excluded data points, resulting in shared learning between the tasks, which is helpful with limited data settings.
Similar to the synthetic dataset, Imputation shows the worst performance both in terms of mean performance as well as variation in the performance. Interestingly, the performance of KMM and Imputation for COVID-19, and MT-Naive, DANN and Imputation for Diabetes falls below Naive method which indicates that the bias handling negatively impacts the performance.
We study the effect of the non-selection rates, event rates and dataset sizes in the following subsections.

\subsubsection{Effect of the Non-selection Rates and the Event Rates}
\label{subsubsec_real_effect_cr}
\begin{figure}[htb!]
     \centering
     \begin{subfigure}[b]{0.32\textwidth}
         \centering
         \includegraphics[width=\textwidth]{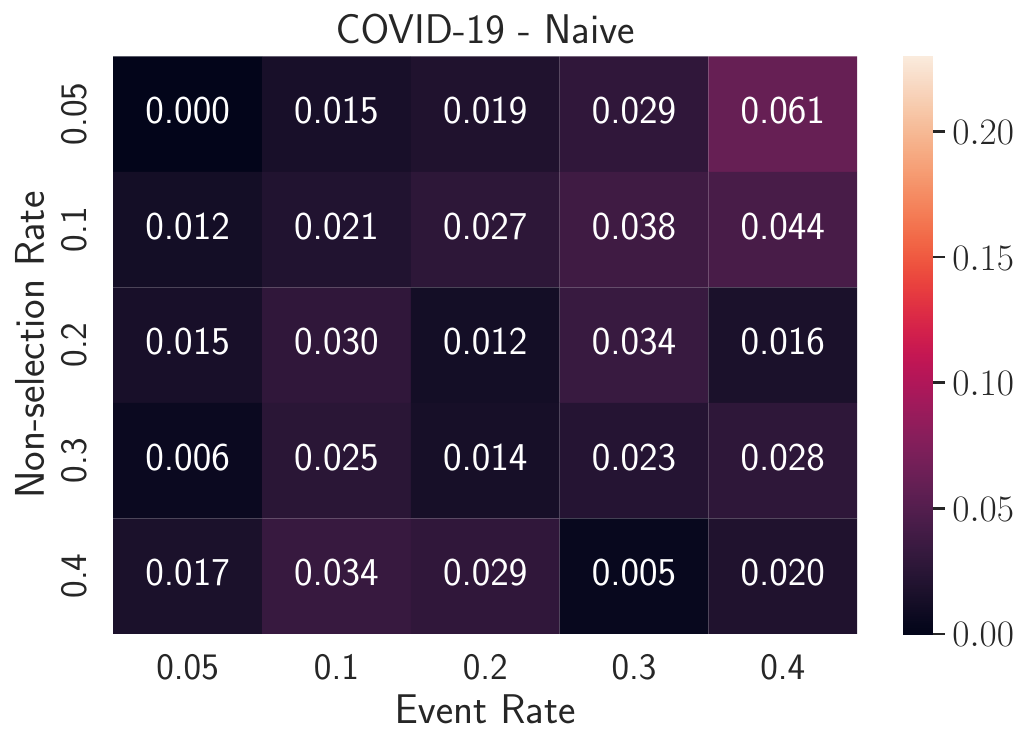}
     \end{subfigure}
    ~
    \begin{subfigure}[b]{0.32\textwidth}
         \centering
         \includegraphics[width=\textwidth]{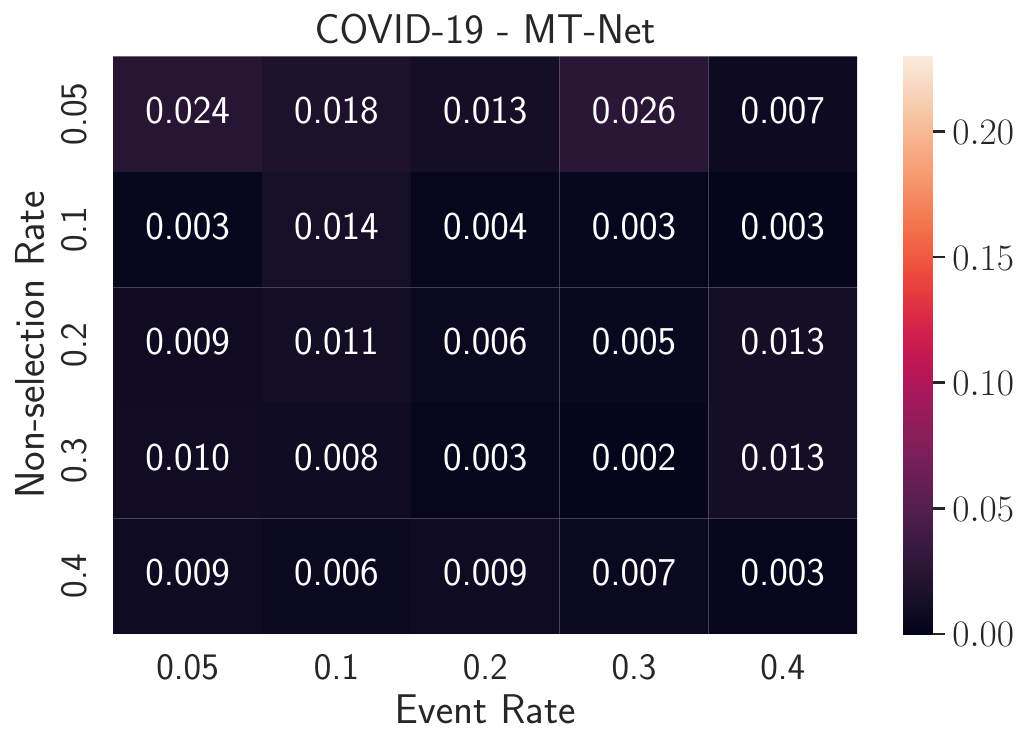}
     \end{subfigure}
     ~
     \begin{subfigure}[b]{0.32\textwidth}
         \centering
         \includegraphics[width=\textwidth]{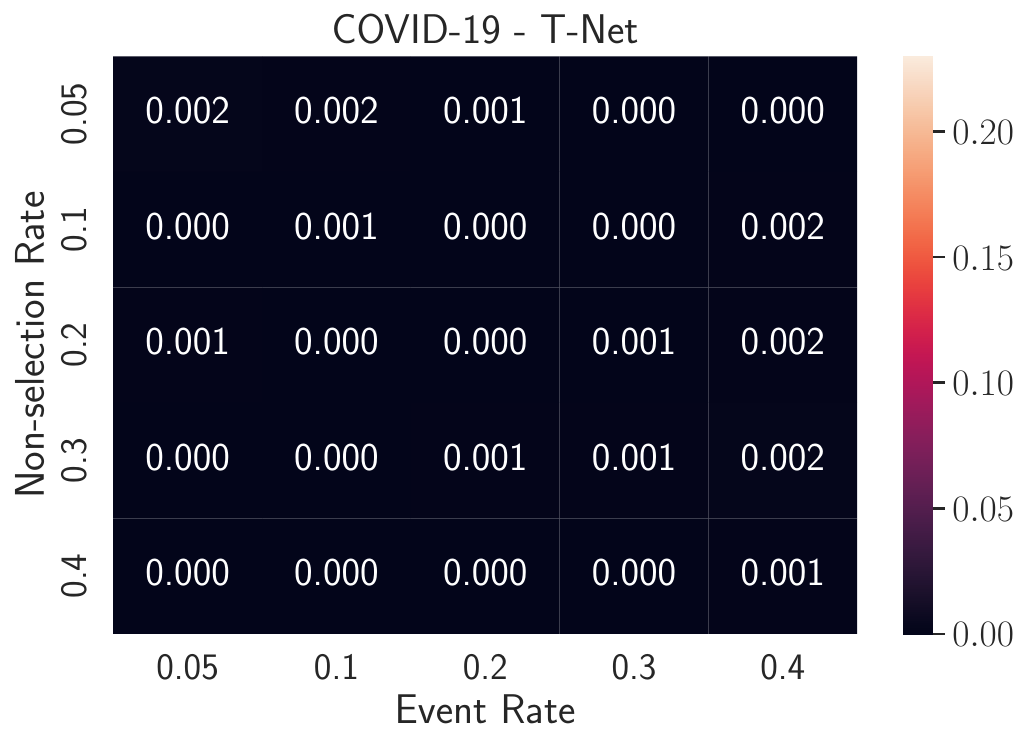}
     \end{subfigure}
     
     \begin{subfigure}[b]{0.32\textwidth}
         \centering
         \includegraphics[width=\textwidth]{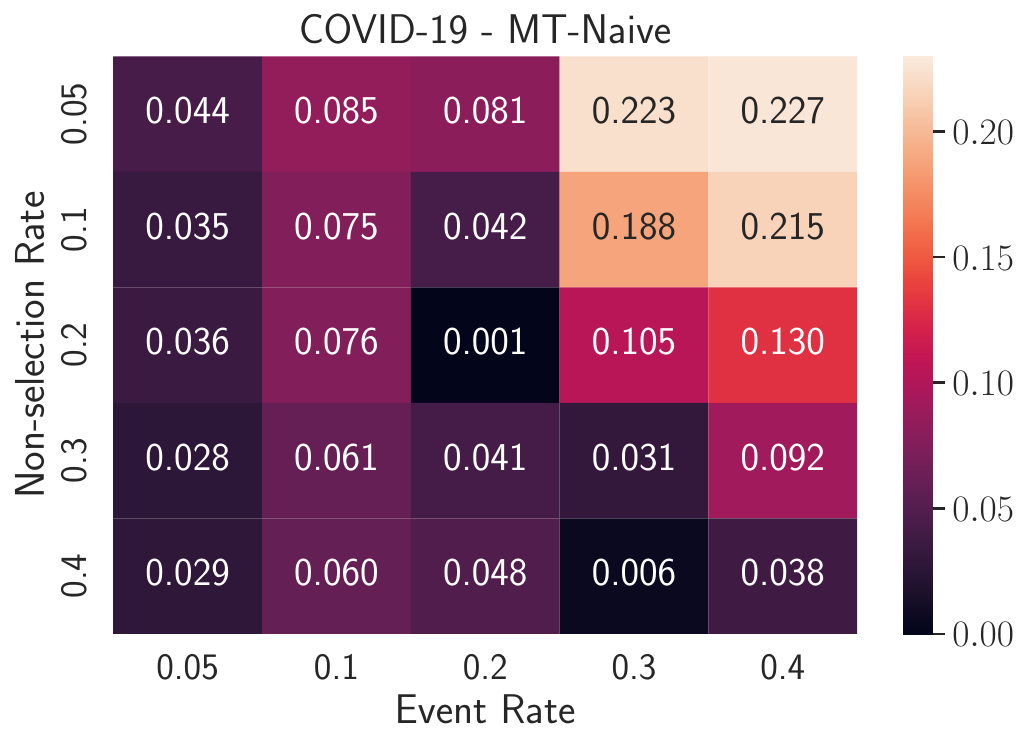}
     \end{subfigure}
     ~
     \begin{subfigure}[b]{0.32\textwidth}
         \centering
         \includegraphics[width=\textwidth]{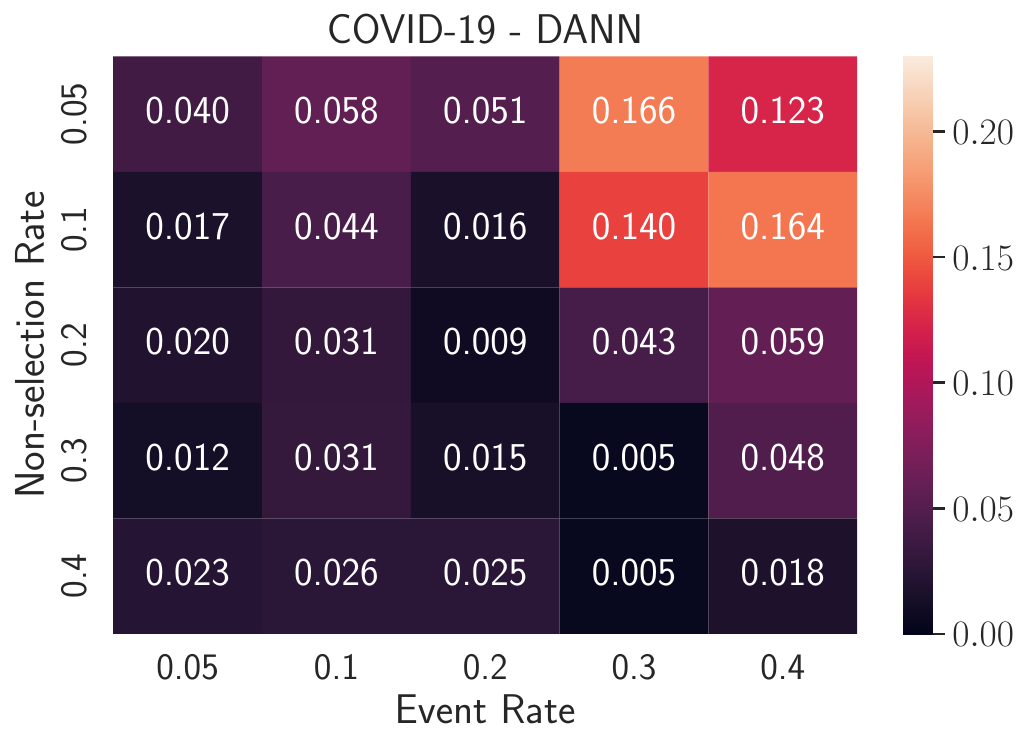}
     \end{subfigure}
     ~
     \begin{subfigure}[b]{0.32\textwidth}
         \centering
         \includegraphics[width=\textwidth]{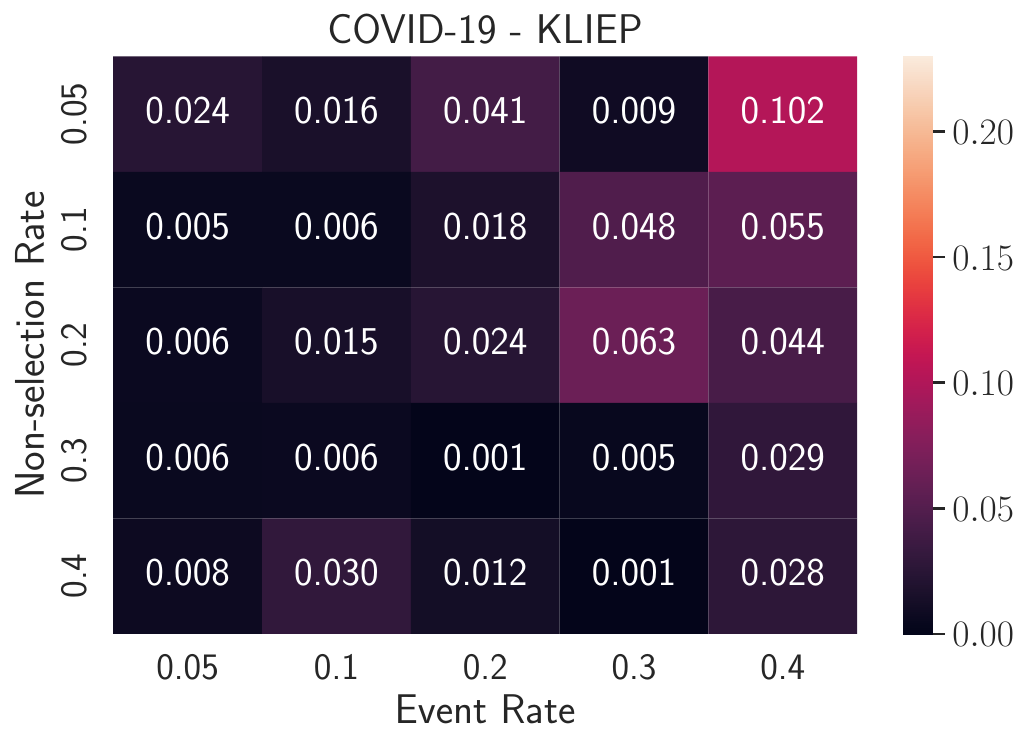}
     \end{subfigure}
    
     \begin{subfigure}[b]{0.32\textwidth}
         \centering
         \includegraphics[width=\textwidth]{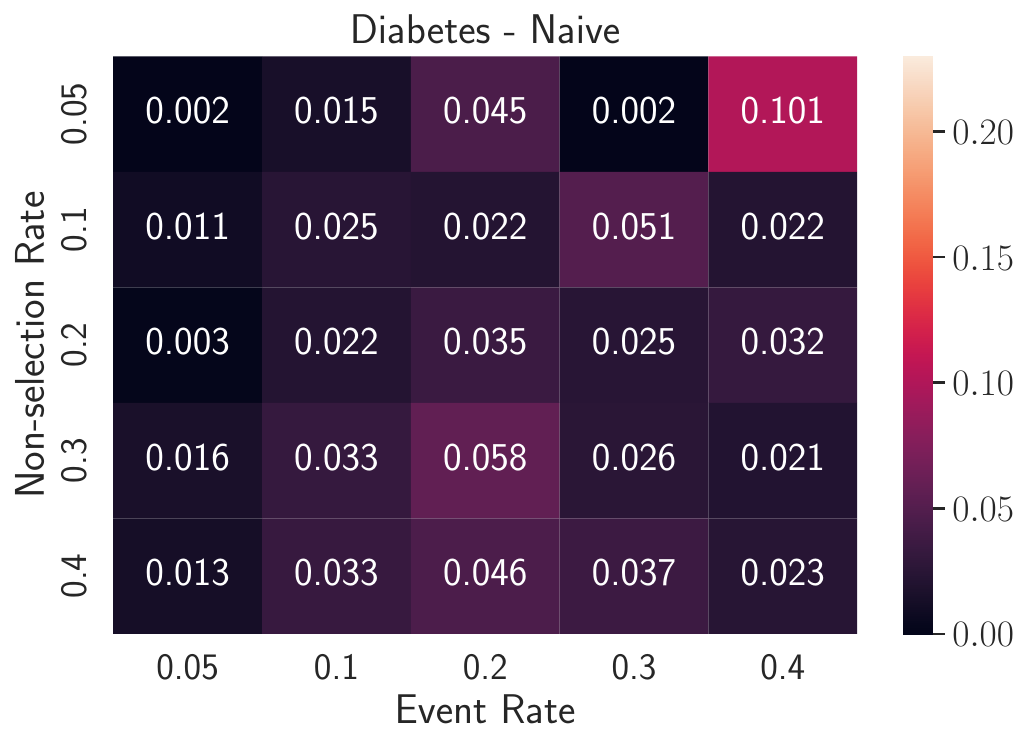}
     \end{subfigure}
     ~
     \begin{subfigure}[b]{0.32\textwidth}
         \centering
         \includegraphics[width=\textwidth]{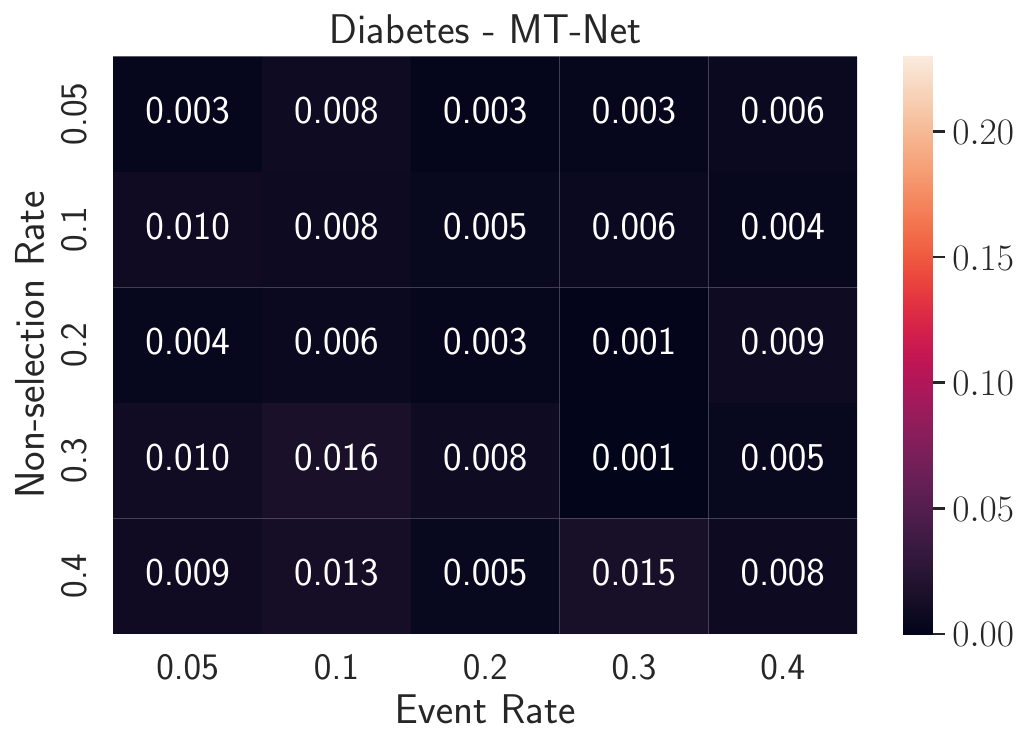}
     \end{subfigure}
     ~
     \begin{subfigure}[b]{0.32\textwidth}
         \centering
         \includegraphics[width=\textwidth]{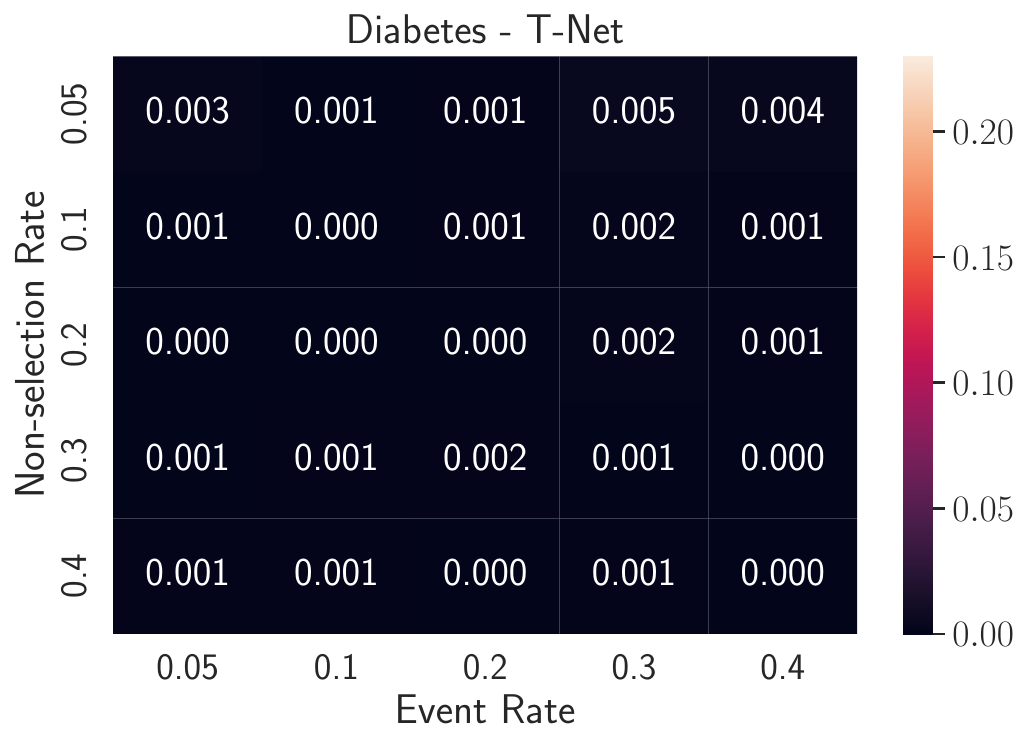}
     \end{subfigure}
    
     \begin{subfigure}[b]{0.32\textwidth}
         \centering
         \includegraphics[width=\textwidth]{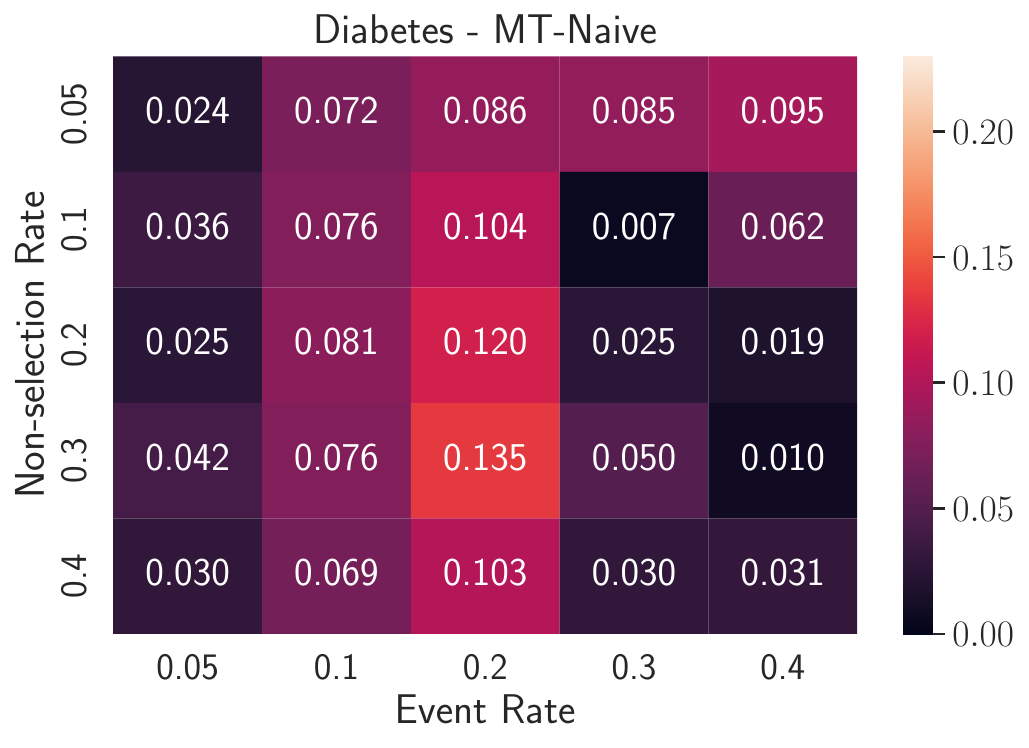}
     \end{subfigure}
     ~
     \begin{subfigure}[b]{0.32\textwidth}
         \centering
         \includegraphics[width=\textwidth]{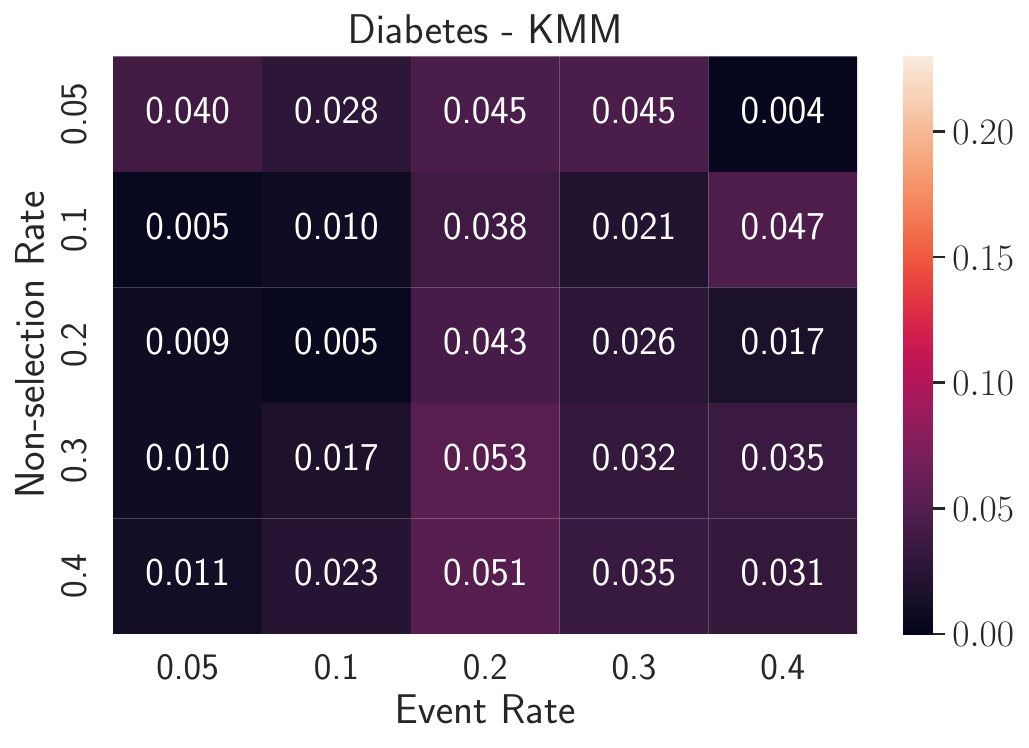}
     \end{subfigure}
     ~
     \begin{subfigure}[b]{0.32\textwidth}
         \centering
         \includegraphics[width=\textwidth]{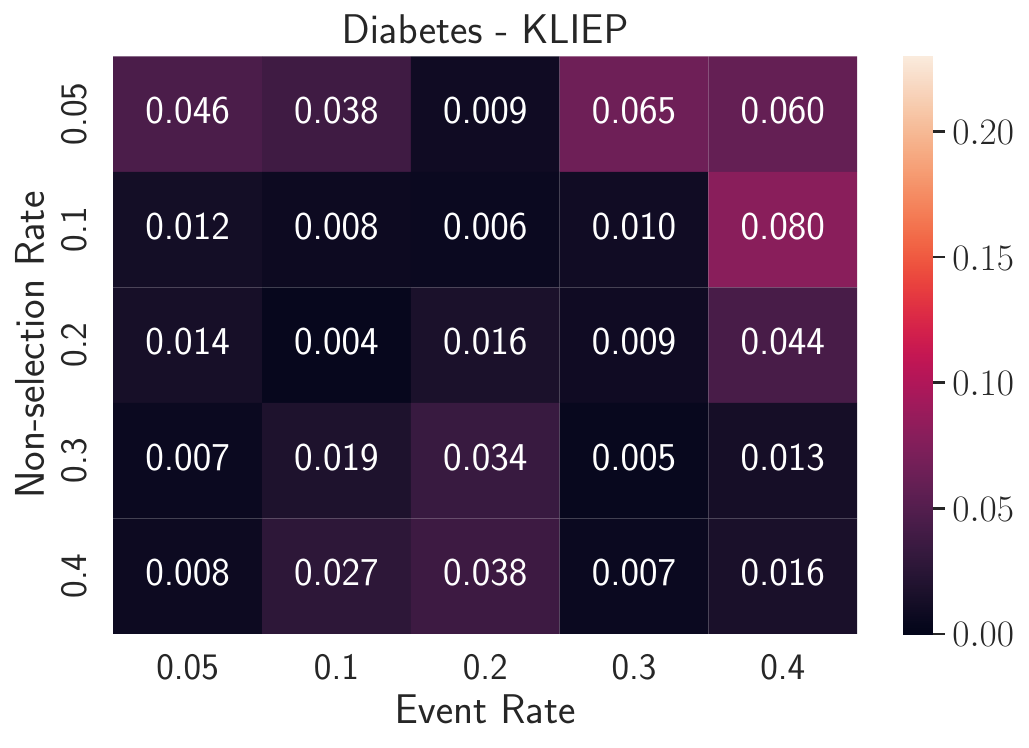}
     \end{subfigure}
     \caption{Effect of event rates and non-selection rates on the performance of different techniques to address SSB using COVID-19 and Diabetes datasets, as averaged over dataset sizes ranging from 1000 to 15,000/25,000.}
     \label{fig_real_risk_vs_selection}
\end{figure}

To study the effect of the non-selection rates and the event rates on the performance of different techniques to address SSB, we consider the event rates and the non-selection rates from [5, 10, 20, 30, 40]\%, and present a summary of the performance averaged over dataset sizes from [1000, 2000, 5000, 10,000, 15,000/25,000] for COVID-19/Diabetes datasets. For the sake of clarity, we consider the proposed techniques and two best baselines, i.e., DANN and KLIEP for COVID-19 and KMM and KLIEP for Diabetes datasets, along with the naive techniques. We present results as absolute differences in their performance from Oracle as heatmaps in Fig.~\ref{fig_real_risk_vs_selection} because Oracle presents an ideal case and unbiased model.

From Fig.~\ref{fig_real_risk_vs_selection}, it is clear that overall MT-Net and T-Net perform better than the baselines as the difference with Oracle is lesser than the baselines.
The relative performance of Naive technique drops more than 6\% for COVID-19 and more than 10\% for Diabetes datasets. As noted earlier, this underlies the importance of addressing SSB in machine learning for healthcare. Similar to the synthetic dataset, overall, the performance difference from Oracle decreases as the event rate increases or the non-selection rate decreases, however, the non-selection rate has more effect on the performance than the event rate.

We observe that MT-Naive and DANN exhibit the highest deviations from Oracle for the COVID-19, with values of 0.227 and 0.166, respectively. For the Diabetes, MT-Naive and KLIEP show the highest deviations, at 0.135 and 0.080, respectively. In contrast, T-Net performs the best across all settings, with maximum deviations of only 0.002 for COVID-19 and 0.005 for Diabetes. Similarly, MT-Net also shows lower deviations as compared to baselines.
Once again, these results show a consistency of the proposed techniques across different settings as compared with the baselines.

\subsubsection{Effect of Dataset Size}
\label{subsubsec_real_effect_scale}
\begin{figure}[htb!]
    \centering
    \includegraphics[width=\linewidth]{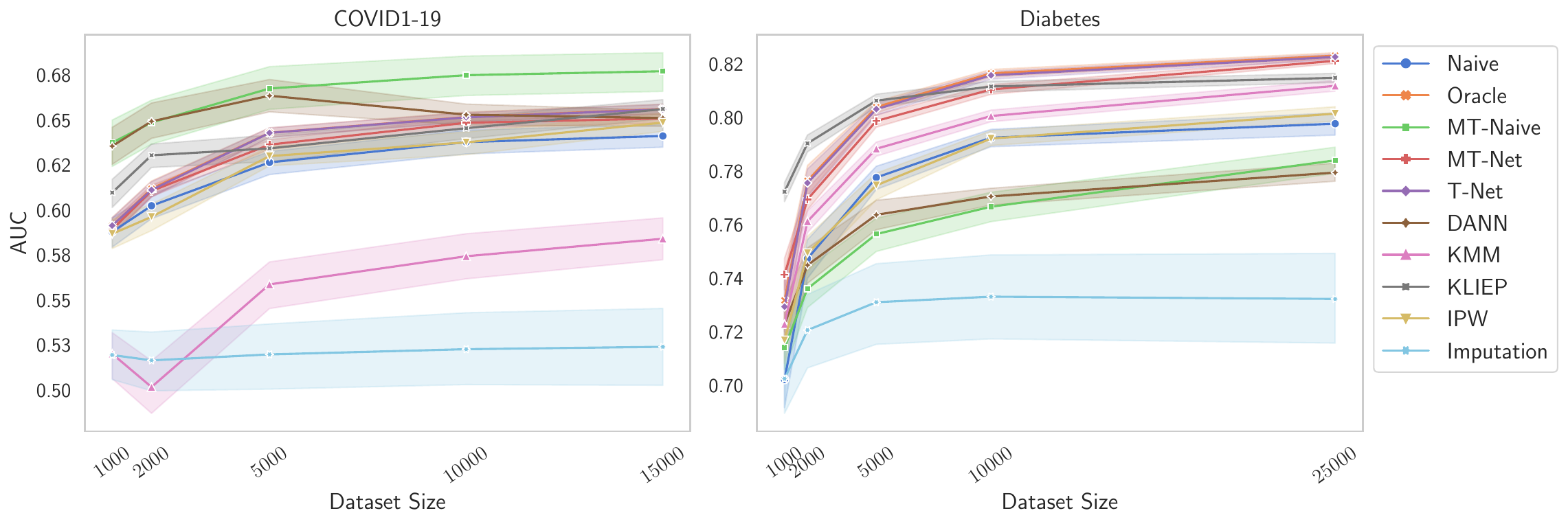}
    \caption{Effect of dataset size on the performance of different techniques to address SSB using COVID-19 and Diabetes datasets, as averaged across event rates and non-selection rates ranging from 5\% to 40\% (shaded region represents one standard deviation over ten runs).}
    \label{fig_real_scale}
\end{figure}

To study the effect of dataset size on the performance of different techniques to address SSB, we consider dataset sizes from [1000, 2000, 3000, 4000, 5000], and present a summary of the performance in Fig.~\ref{fig_real_scale} averaged over the event rates and the non-selection rates from [5, 10, 20, 30, 40]\%.

From the figure, we observe that MT-Net and T-Net techniques perform close to Oracle, which is the desired behaviour when we are addressing SSB in the dataset. Similar to the synthetic dataset, Imputation performs the worst for both the datasets, and Oracle performs the best on Diabetes dataset but for COVID-19 dataset MT-Naive performs the best.
For COVID-19 dataset, KMM and Imputation perform worse than Naive, while for Diabetes dataset, MT-Naive, Imputation and DANN perform worse than Naive. This indicates the bias correction techniques fail and negatively impact the performance.

As observed earlier, MT-Net is more effective than T-Net for smaller datasets due to access to included and excluded patients and from sharing of information. The proposed techniques perform consistently across settings and datasets, while the baselines show huge fluctuations, for example, MT-Naive outperforms the rest for COVID-19 but falls even below Naive for Diabetes. Similarly, KLIEP shows good performance for smaller datasets as observed for COVID-19 and Diabetes but lagged for the synthetic dataset.
Overall, as expected, the performance of all the techniques for addressing SSB improves with an increase in the dataset size.

\subsection{Effect of SSB on the Non-selected Subpopulation}
\label{subsec_effect_non_selected}
\begin{figure}[htb!]
    \centering
    \includegraphics[width=0.7\linewidth]{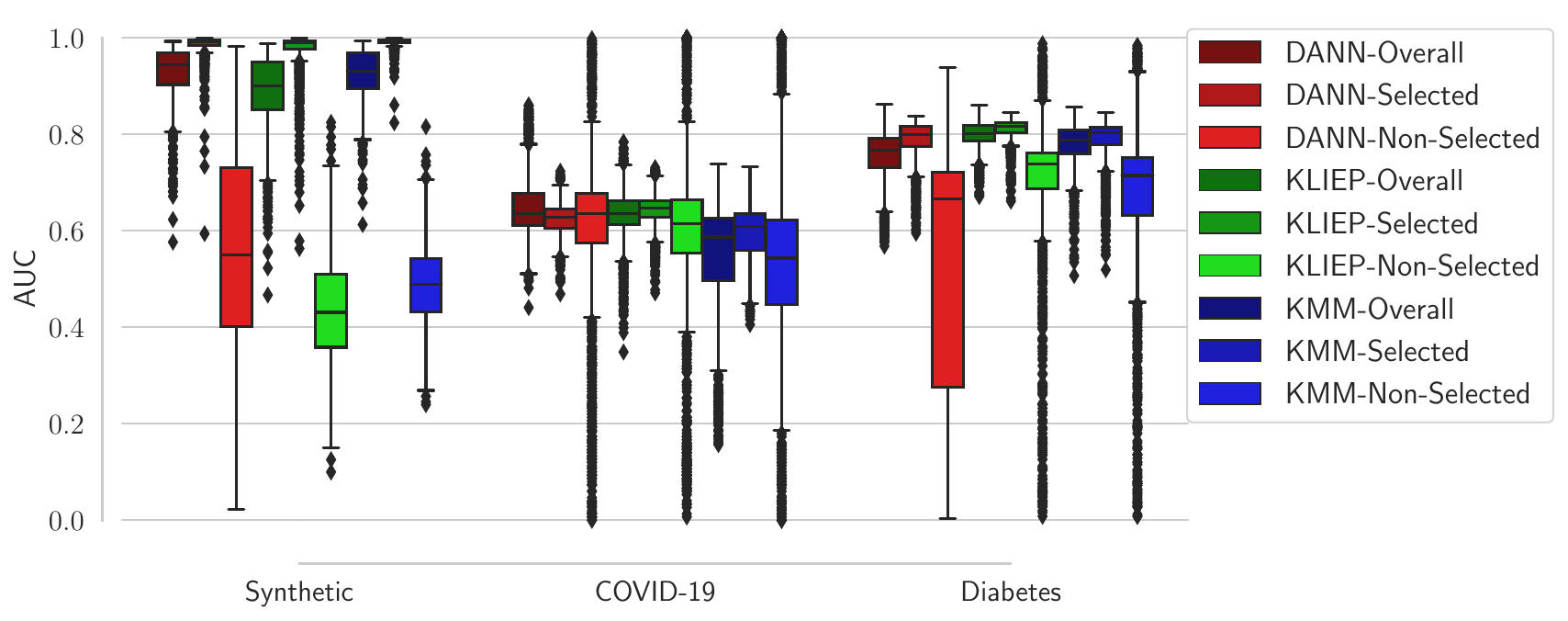}
    \caption{Analysis of the average performance for the selected, the non-selected and the overall population for the best baselines using the synthetic, COVID-19 and Diabetes datasets averaged over the event rates and the non-selection rates ranging from 5\% to 40\%, and dataset sizes ranging from 1000 to 5000/15,000/25,000.}
    \label{fig_included_vs_excluded}
\end{figure}

We analyse the performance of the best SSB handling baselines to study their performance for the selected, the non-selected and all the patients of the study with the synthetic and semi-synthetic datasets, and present results in Fig.~\ref{fig_included_vs_excluded}\footnote{We can't perform such an analysis for the proposed techniques as they make predictions only for the selected subpopulation.}.
From the figure, we observe a large difference in the performance of the SSB handling techniques between the selected subpopulation compared to the non-selected subpopulation. We observe the largest difference in the performance for the synthetic dataset where the mean difference in the performance is 43\%, 56\% and 51\% for DANN, KLIEP and KMM, respectively. For COVID-19, the mean performance difference is 5\%, 8\% and 7\% for DANN, KLIEP and KMM, respectively, and for Diabetes, the mean performance difference is 33\%, 15\% and 16\% for DANN, KLIEP and KMM, respectively.
For the synthetic dataset, on one side the SSB handling techniques show near perfect performance for the patients selected in the study population but on the other side for the non-selected subpopulation, these techniques show very poor performance, close to 50\% which show random predictions for the non-selected patient population. 
For the synthetic dataset, the performance for the non-selected subpopulation is always lesser than the included subpopulation, however for the semi-synthetic datasets, there is a larger variation in the performance compared to the synthetic dataset.
We did not find any consistent pattern in the performance of the baselines which could show good performance for the non-selected subpopulation as well as the selected subpopulation. DANN shows the best performance for the synthetic and COVID-19 datasets but performs the worst for Diabetes dataset.
Thus, the baselines lead to a poor performance for the non-selected subpopulation, resulting in a strong bias, for the non-selected subpopulation compared to the selected subpopulation. This would lead to harmful decisions for the non-selected subpopulation. Thus, these results highlight the dangers and the need to address SSB in machine learning algorithms for healthcare.

\subsection{Ablation of TPI Approach}
\label{subsec_results_identification}
\begin{figure}[htb!]
     \centering
     \begin{subfigure}[b]{0.32\textwidth}
         \centering
         \includegraphics[width=\textwidth]{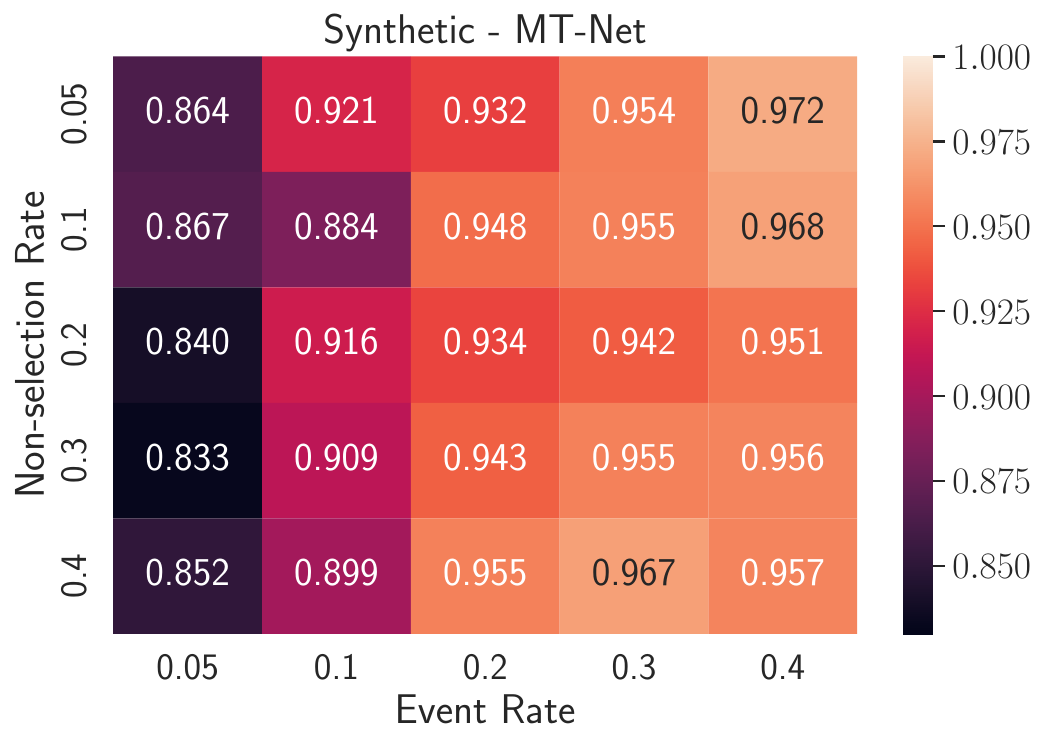}
     \end{subfigure}
     ~
     \begin{subfigure}[b]{0.32\textwidth}
         \centering
         \includegraphics[width=\textwidth]{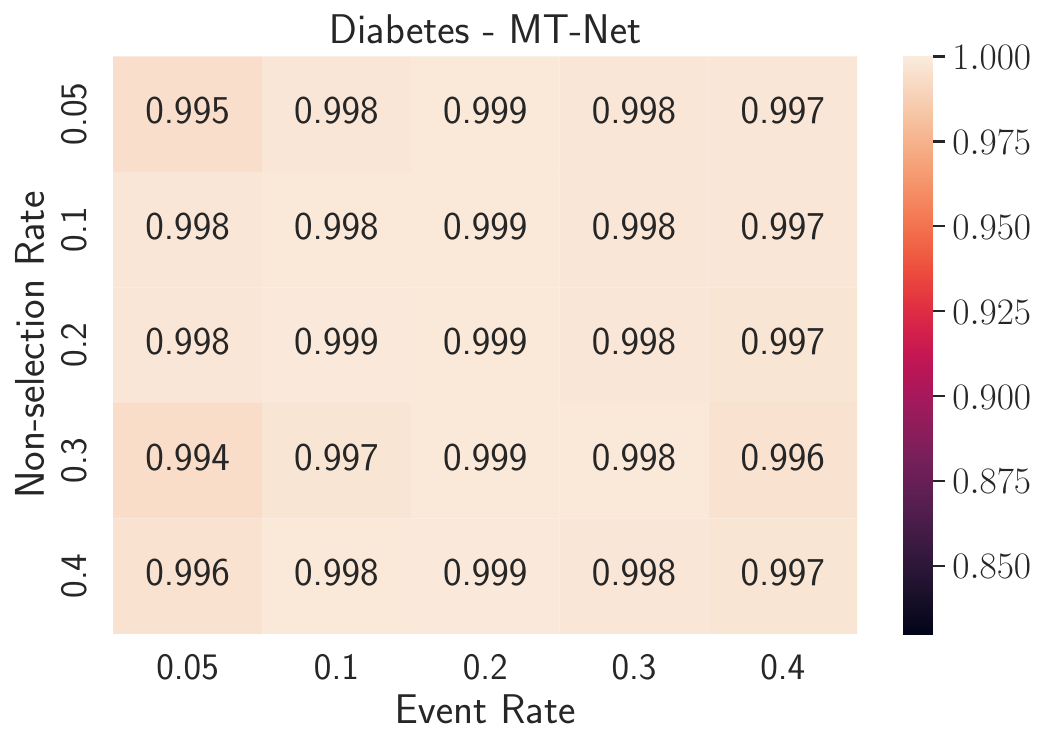}
     \end{subfigure}
     ~
     \begin{subfigure}[b]{0.32\textwidth}
         \centering
         \includegraphics[width=\textwidth]{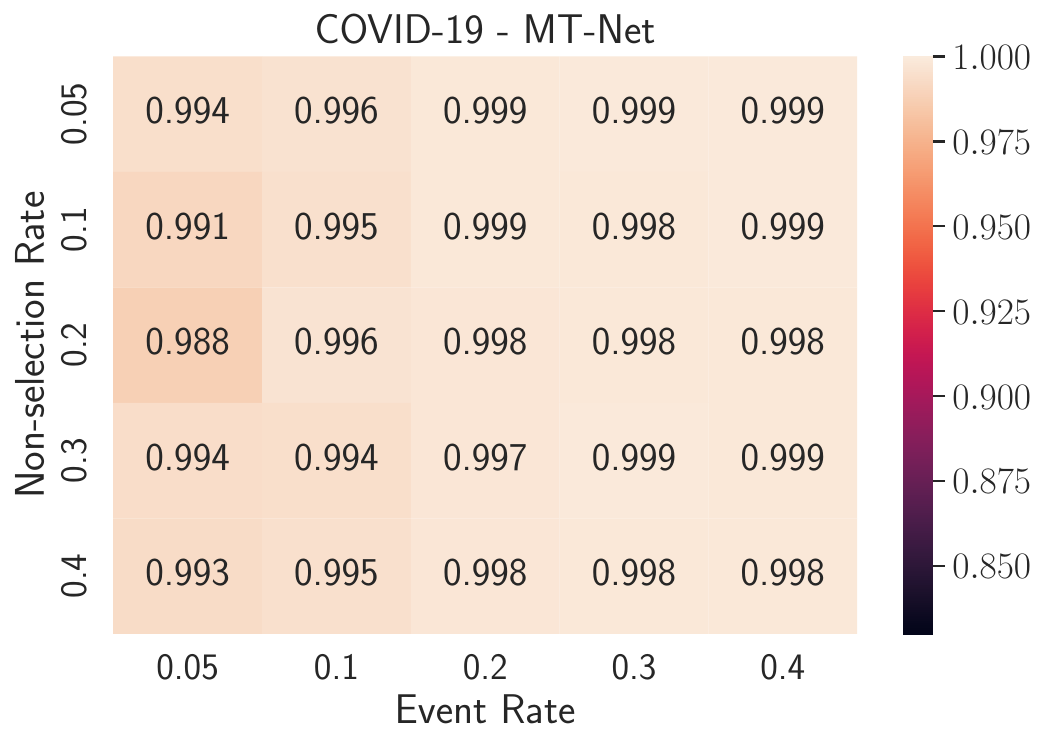}
     \end{subfigure}
    
     \begin{subfigure}[b]{0.32\textwidth}
         \centering
         \includegraphics[width=\textwidth]{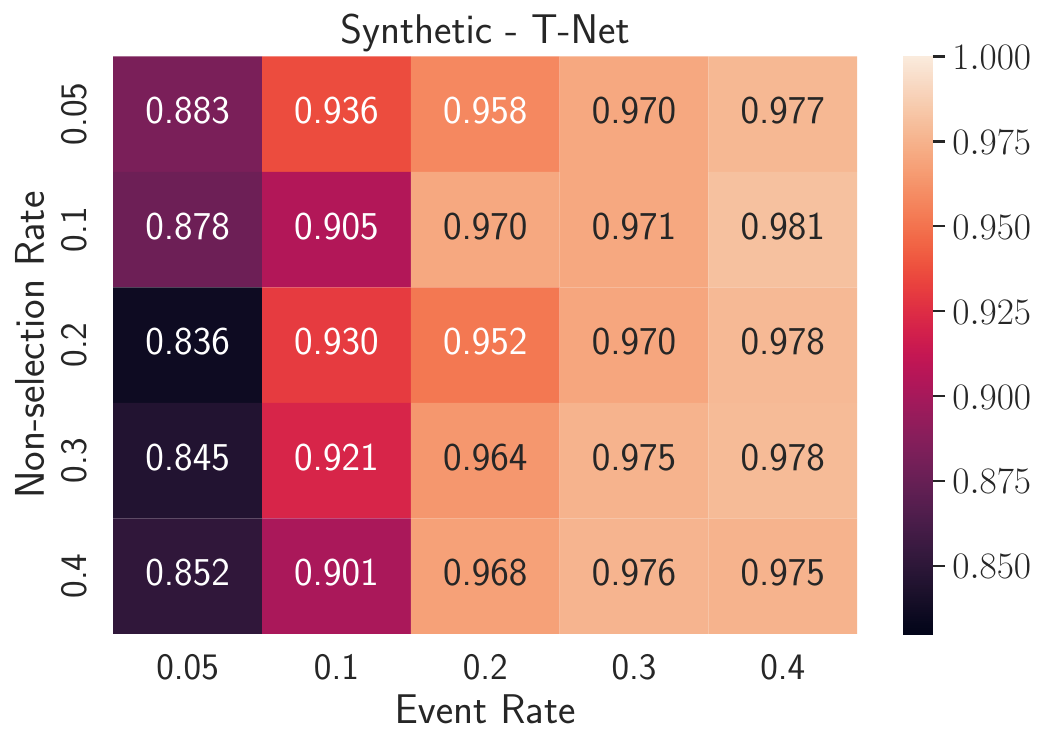}
     \end{subfigure}
    ~
     \begin{subfigure}[b]{0.32\textwidth}
         \centering
         \includegraphics[width=\textwidth]{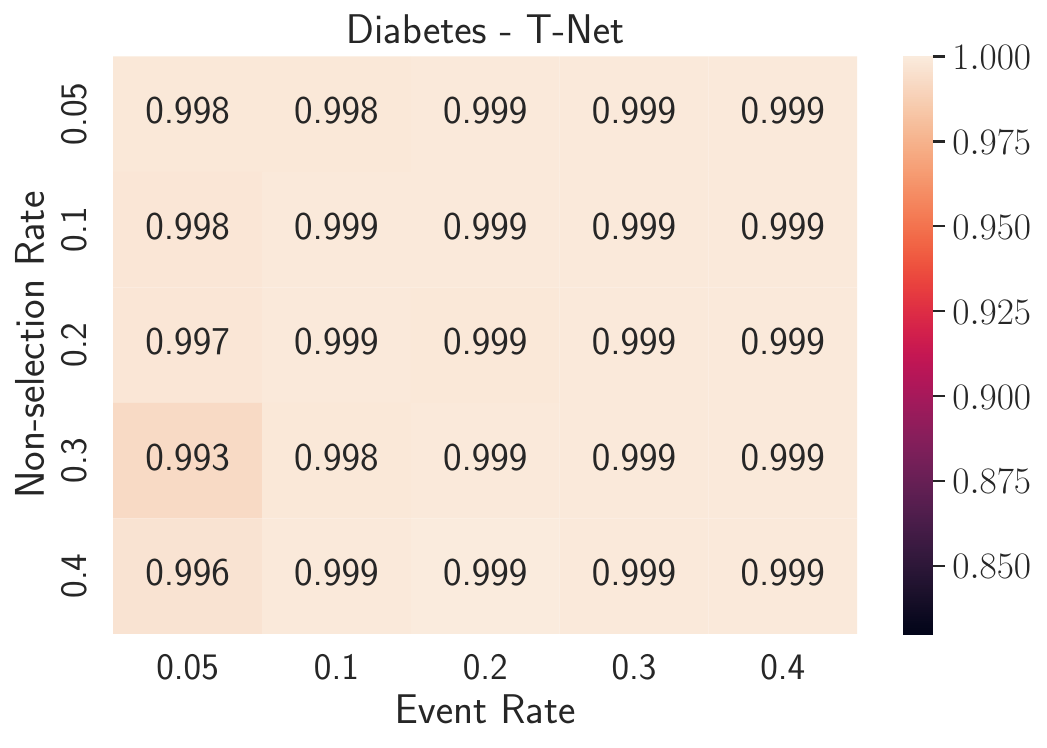}
     \end{subfigure}
    ~
     \begin{subfigure}[b]{0.32\textwidth}
         \centering
         \includegraphics[width=\textwidth]{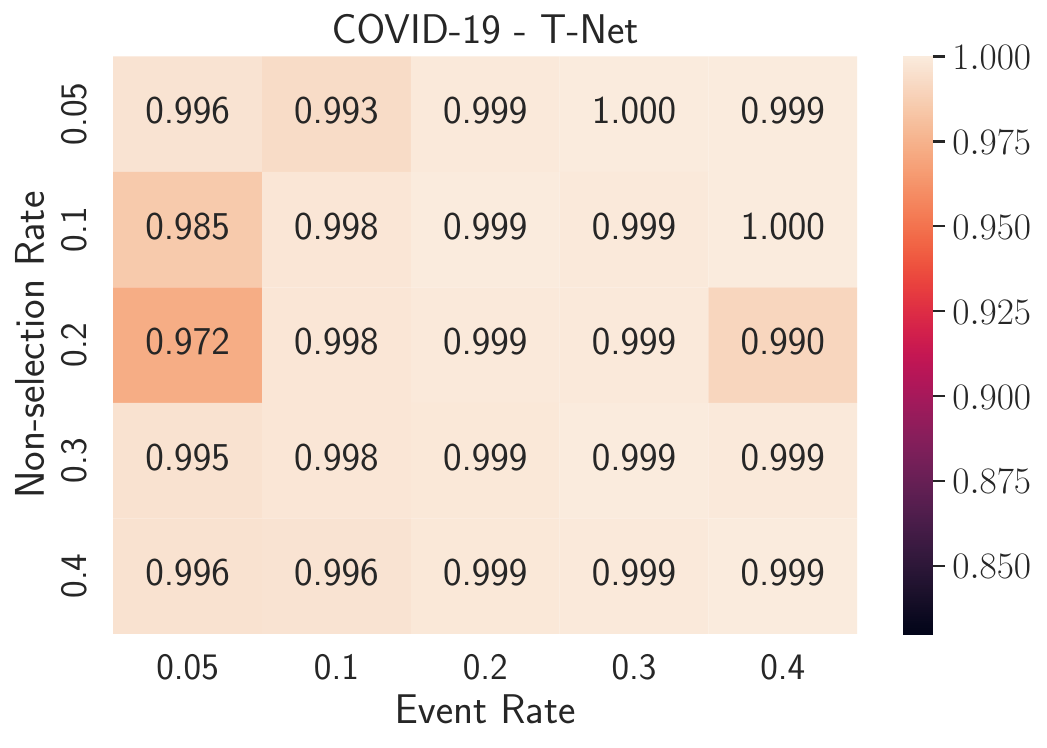}
     \end{subfigure}
     \caption{Analysis of the average performance of MT-Net and T-Net in AUC to identify the target population with the synthetic and the semi-synthetic datasets.}
     \label{fig_real_risk_vs_selection-identification}
\end{figure}

Here, we analyse MT-Net and T-Net's ability to identify the target population with the synthetic, COVID-19 and Diabetes datasets, measured using AUC. The results are presented in Fig.~\ref{fig_real_risk_vs_selection-identification}.
We observe that overall MT-Net and T-Net techniques show good performance in identifying the target population. They show better performance on the semi-synthetic datasets as the performance drops from one to three percent from the perfect score, while they show a drop of three to 15\% from the perfect score for the synthetic dataset. This is potentially due to the large dataset sizes of the semi-synthetic datasets and using different mechanisms for data generation.
This figure also explains the overall pattern in the performance of the proposed techniques as their performance drops as their ability to identify the target population drops.
In general, we observe that the performance improves with a decrease in the non-selection rate or an increase in the event rate. However, the effect of the event rate is more pronounced as compared to the non-selection rate. The effect of the non-selection rate is a bit unexpected, however, the effect of the event rate is as expected because, with an increase in the event rate, the class imbalance decreases.
MT-Net and T-Net show similar performance for different studied settings.

\section{Discussion}
\label{sec_discussion}
SSB is a result of a non-random process to select patients in the study population while employing a random process to select patients in the target population, resulting in a distribution shift between the study population and the target population. The algorithms developed with the biased study population could lead to inaccurate predictions and harmful decisions for patients not selected in the study population. Despite being well known for decades and a threat to the validity of results, SSB is scarcely addressed in machine learning for healthcare. On the other hand, the existing techniques to address SSB, developed primarily in machine learning, are based on the bias correction approach where distributions of the study population and the target population are balanced to be similar. The existing balancing-based approach leads to the loss of predictive power of the algorithms, resulting in a loss of performance. Thus, the main objective of our work is to highlight the potential dangers of SSB, if unaddressed, on the performance of machine learning algorithms, and propose a new research direction for addressing the limitations of the existing techniques.

Our literature review and empirical study results converge on a crucial point that the presence of SSB in machine learning for healthcare poses significant risks to algorithmic performance and fairness. Several reviews have highlighted this issue, emphasising the potential harms associated with unaddressed SSB. For example, a meta-review of 50 reviews consisting of 2104 models by \citet{Jong2021appraising} found that information on the risk of bias per domain was available for 1039 (47\%) studies, of which 25\% have a high risk of bias in participant selection. Our analysis of synthetic and semi-synthetic datasets aligns with these concerns. We observe that SSB could lead to a large overall performance drop of up to 22\% in machine learning algorithms and a substantial performance gap of more than 40\% between the non-selected subpopulation and the selected subpopulation, showing inaccurate and biased predictions for the non-selected subpopulation. These findings highlight the critical importance of addressing SSB in machine learning algorithms for healthcare.

The existing techniques, borrowed from other machine learning areas like domain-adaption, are mostly based on the idea of bias correction and try to match distributions between the study population and the target population, leading to a loss of predictive performance of the machine learning algorithms. To address this issue, we propose a new research direction to address SSB based on the TPI, rather than bias correction. Here, an algorithm would focus first on identifying the target subpopulation representative of the study population and seen by the algorithm during the training time, and then making predictions for the identified subpopulation, while deferring the rest of the patients to clinicians.
Thus, the proposed approach avoids the loss of predictive power due to alignment of the study and the target population distributions as is reflected by consistent superior performance across settings, including varying non-selection rates, event rates, dataset sizes as well as across multiple datasets. 
This approach bears resemblance to uncertainty quantification where predictions are generated only for those patients where an algorithm is confident otherwise the patients are referred to a clinician \cite{kompa2021second,chauhan2024continuous} which is essential for the safe application of machine learning \cite{guo2017calibration,minderer2021revisiting}. Uncertainty quantification and TPI can also be regarded as subcategories within the broader framework of machine learning with rejection, where algorithms refrain from making predictions when they are likely to err \cite{hendrickx2024machine}. The proposed approach is motivated by healthcare where certain patient populations fail to complete a study and the target population would contain any patients, including patient groups who failed to complete the study. This setting is unlike domain adaptation where the target dataset is supposed to be different from the source and the algorithms are supposed to work with the target dataset. Additionally, compared to the naive algorithms, the proposed approach can utilise the non-selected patients for training machine learning algorithms which could be helpful for limited data settings.

We propose two concrete techniques under the proposed TPI approach, called T-Net and MT-Net, which use two independent networks and a multitasking approach, respectively. They develop one network/task for predicting the selection of a patient in the study population and a second network/task for the risk prediction task. T-Net has the flexibility to be more expressive than MT-Net as it develops two independent neural networks for the selection and the risk prediction task. On the other hand, MT-Net has the advantage of information sharing across the two tasks due to shared learning which benefits from inductive learning. As our analysis finds, these properties of the proposed techniques make them suitable for certain settings, for example, MT-Net is better suited for limited data settings and settings with low non-selection rates, while T-Net is better suited for the rest of the settings.

Our analysis of the synthetic and the semi-synthetic datasets shows that the proposed SSB handling techniques are very effective across all settings and show consistent results for different selection rates, event rates, dataset sizes and datasets. The proposed techniques always improve over Naive technique while the baselines sometimes fall below Naive, indicating a negative effect of bias correction. For limited data settings, MT-Net is more effective than T-Net.

It is interesting to note that SSB might not always lead to inaccurate predictions for the non-selected population. This is because SSB impacts the performance of machine learning algorithms only when the non-selected patients play an active role in deciding the decision boundary and unequally favour different outcomes. Suppose, the non-selected patients lie away from the decision boundary but on the correct side then there will be no impact of SSB on the performance of algorithms, even when trained on the biased dataset. This means that by definition, Naive method will perform better than all the baselines because the predictions will be correct for the non-selected patients, resulting in better performance for Naive method. Moreover, all bias handling techniques would lead to a loss of performance. Thus, in such a setting, Naive method, i.e., ignoring the bias would be the best technique to deal with SSB. We can easily identify such a setting, and suggest the community to always develop Naive method, in addition to SSB handling techniques, and if Naive shows improvement in its performance on the target population as compared with the study population then that is an indicator that SSB does not affect the decision boundary and has no impact on algorithms.

\textbf{Limitations and Future Work:} The proposed TPI approach operates under the assumption, akin to IPW and imputation methods, that we can infer a patient's selection or non-selection into a study population based on their attributes. Essentially, we assume a missing-at-random mechanism for outcomes of the non-selected patients, a common practice in statistical and clinical studies employing imputation techniques to address SSB. Moreover, similar to IPW, the proposed techniques work in two steps, where the first step identifies the target subpopulation representative of the study population, followed by predictions for the identified subpopulation. So, any misspecification in the first step would lead to errors in the second step. However, since the identification task predicts the selection of patients for inclusion in the study, it has access to patients included as well as excluded in the study, and hence relative to the prediction task it has more access to data that would reduce errors in learning.

The proposed techniques make predictions only for the target subpopulation representative of the study population, i.e., the selected patients. However, we argue that non-random sample selection is inevitable in healthcare. For example, clinical studies often have strict inclusion and exclusion criteria (e.g., only adult patients), meaning the results apply to the restricted population (only adult patients). Therefore, as long as the study and target population align, bias does not occur, and the algorithms remain useful. It is better to make accurate predictions for subpopulations than to make biased predictions for certain subpopulations when considering the entire population. Thus, our SSB handling approach ensures robust and reliable predictions within the defined subpopulation, contributing to more accurate and effective healthcare decision-making.

The proposed techniques for selective predictions may raise ethical concerns if the non-represented patients in the study population belong to marginalised subgroups, thus exacerbating existing social and health inequalities. There are three potential approaches to addressing this issue, as discussed in \cite{vandersluis2024selective,goetz2024generalization}: (i) not using algorithms at all – which also presents an ethical dilemma, as not using algorithms could harm subpopulations that might benefit from their application, (ii) employing bias-correction algorithms, which risk making harmful decisions for non-represented subpopulations, and (iii) using selective predictions and referring non-selected patients to clinicians. While both options one and two present ethical concerns, option three – selective predictions – is a potentially viable approach emerging in fields like bioethics \cite{vandersluis2024selective}. By referring non-represented subpopulations to clinicians, their care is not negatively affected, while selective predictions could benefit certain subpopulations. While our focus is on developing methodology, we believe that the use of selective predictions should be context-dependent and involve input from the relevant stakeholders.

Another limitation of the proposed work is that it is limited to setting where the target dataset is a superset of the source dataset, and may not work for domain adaptation where the entire target dataset could be different from the source. Hence, it would be interesting to explore settings across disciplines where the proposed approach could be beneficial and will be explored as part of our future work.
Validating SSB handling techniques with real-world datasets poses a challenge, as outcomes for patients excluded from the study are unavailable. However, we can evaluate the effectiveness of these techniques by comparing their performance on the target population against the naive machine learning algorithm on the study population. A smaller performance gap indicates better handling of SSB. It's worth noting that other biases or noise in the target population could contribute to performance differences. Nonetheless, this comparative analysis will offer insight into the relative effectiveness of different techniques. Future research will focus on exploring and validating our proposed methodology with real-world healthcare studies.

\section{Conclusion}
\label{sec_conclusion}
This paper underscores the critical need to address SSB in machine learning for healthcare, highlighting its potential to compromise algorithm performance and lead to inappropriate interventions for unrepresented subpopulations.
To tackle this issue, we propose a novel research direction focusing on identifying the target subpopulation representative of the study population and making predictions specifically for these identified patients. In contrast to existing techniques that focus on correcting the bias, our proposed approach also prevents the loss of predictive accuracy caused by aligning distributions between the study and target populations.
Furthermore, we introduce T-Net and MT-Net as implementations of the TPI approach. These models are trained to predict both the underlying task and the selection of patients into the study population. Our empirical results demonstrate that SSB leads to a large drop in performance for the target population compared to the study population and a substantial performance difference for the target subpopulations representative of the selected and the non-selected patients. Additionally, T-Net and MT-Net exhibit consistent performance in addressing SSB across various scenarios, including different non-selection rates, event rates, dataset sizes, and datasets.

\section*{Competing Interests}
The authors declare that they have no competing interests.

\section*{Availability of data and materials}
The data used in the study are publicly available from the following links: Diabetes dataset is released by the Center for Disease Control and Prevention and available on Kaggle at \url{https://www.kaggle.com/datasets/cdc/behavioral-risk-factor-surveillance-system}. A cleaned, consolidated and balanced version is available at \url{https://www.kaggle.com/datasets/alexteboul/diabetes-health-indicators-dataset}. COVID-19 dataset is an anonymised dataset released by the Mexican government \url{https://www.gob.mx/salud/documentos/datos-abiertos-152127} and accessed from Kaggle \url{https://www.kaggle.com/datasets/tanmoyx/covid19-patient-precondition-dataset/data}.

\section*{Availability of code}
The complete code is available at \url{https://github.com/jmdvinodjmd/SSB_MLHC}.

\section*{Acknowledgements}
This work was supported in part by the National Institute for Health Research (NIHR) Oxford Biomedical Research Centre (BRC) and in part by the ITC InnoHK ``Oxford-CityU Hong Kong Centre for Cerebrocardiovascular Health Engineering'' (COCHE). DAC was supported by an NIHR Research Professorship, an RAEng Research Chair, and the Pandemic Sciences Institute at the University of Oxford. GN is funded by the NIHR (Grant number 302607) for a doctoral research fellowship. 
The views expressed are those of the authors and not necessarily those of the NHS, the NIHR, the Department of Health, the InnoHK – ITC, or the University of Oxford.

We also thank the reviewers for their insightful feedback, which has greatly improved the quality of this work.


\bibliographystyle{apalike}
\bibliography{ML4HC}

\begin{thebibliography}{}

\bibitem[Alaimo Di~Loro et~al., 2023]{Alaimo2023}
Alaimo Di~Loro, P., Scacciatelli, D., and Tagliaferri, G. (2023).
\newblock 2-step gradient boosting approach to selectivity bias correction in
  tax audit: an application to the vat gap in italy.
\newblock {\em Statistical Methods \& Applications}, 32(1):237--270.

\bibitem[Banack et~al., 2019]{banack2019investigating}
Banack, H.~R., Kaufman, J.~S., Wactawski-Wende, J., Troen, B.~R., and Stovitz,
  S.~D. (2019).
\newblock Investigating and remediating selection bias in geriatrics research:
  the selection bias toolkit.
\newblock {\em Journal of the American Geriatrics Society}, 67(9):1970--1976.

\bibitem[Berk, 1983]{berk1983introduction}
Berk, R.~A. (1983).
\newblock An introduction to sample selection bias in sociological data.
\newblock {\em American sociological review}, pages 386--398.

\bibitem[Bia et~al., 2023]{bia2023double}
Bia, M., Huber, M., and Laff{\'e}rs, L. (2023).
\newblock Double machine learning for sample selection models.
\newblock {\em Journal of Business \& Economic Statistics}, pages 1--12.

\bibitem[Bishop, 1995]{bishop1995neural}
Bishop, C. (1995).
\newblock Neural networks for pattern recognition.
\newblock {\em Clarendon Press google schola}, 2:223--228.

\bibitem[Bonita et~al., 2006]{bonita2006basic}
Bonita, R., Beaglehole, R., and Kjellstr{\"o}m, T. (2006).
\newblock {\em Basic epidemiology}.
\newblock World Health Organization.

\bibitem[Boonstra et~al., 2021]{boonstra2021simulation}
Boonstra, P.~S., Little, R.~J., West, B.~T., Andridge, R.~R., and
  Alvarado-Leiton, F. (2021).
\newblock A simulation study of diagnostics for selection bias.
\newblock {\em Journal of official statistics}, 37(3):751--769.

\bibitem[Bradley and Nichols, 2022]{bradley2022addressing}
Bradley, V. and Nichols, T.~E. (2022).
\newblock Addressing selection bias in the uk biobank neurological imaging
  cohort.
\newblock {\em MedRxiv}, pages 2022--01.

\bibitem[Breen et~al., 2015]{Breen_2015}
Breen, R., Choi, S., and Holm, A. (2015).
\newblock Heterogeneous causal effects and sample selection bias.
\newblock {\em Sociological Science}, 2(17):351--369.

\bibitem[Brewer et~al., 2021]{brewer2021addressing}
Brewer, D., Carlson, A., et~al. (2021).
\newblock Addressing sample selection bias for machine learning methods.
\newblock Technical report, Department of Economics, University of Missouri.

\bibitem[Chauhan et~al., 2025]{Chauhan2025beyond}
Chauhan, V.~K., Dhami, D.~S., Gao, B., Wang, X., Clifton, L., and Clifton,
  D.~A. (2025).
\newblock {Beyond Correlations: The Necessity and the Challenges of Causal AI}.
\newblock {\em TechRxiv,
  http://dx.doi.org/10.36227/techrxiv.175554759.96327720/v1}.

\bibitem[Chauhan et~al., 2023]{chauhan2023adversarial}
Chauhan, V.~K., Molaei, S., Tania, M.~H., Thakur, A., Zhu, T., and Clifton,
  D.~A. (2023).
\newblock Adversarial de-confounding in individualised treatment effects
  estimation.
\newblock In {\em Proceedings of The 26th International Conference on
  Artificial Intelligence and Statistics}, volume 206, pages 837--849. PMLR.

\bibitem[Chauhan et~al., 2024a]{chauhan2024continuous}
Chauhan, V.~K., Thakur, A., O’Donoghue, O., Rohanian, O., Molaei, S., and
  Clifton, D.~A. (2024a).
\newblock Continuous patient state attention model for addressing irregularity
  in electronic health records.
\newblock {\em BMC Medical Informatics and Decision Making}, 24(1):117.

\bibitem[Chauhan et~al., 2024b]{chauhan2024dynamic}
Chauhan, V.~K., Zhou, J., Ghosheh, G., Molaei, S., and A~Clifton, D. (2024b).
\newblock Dynamic inter-treatment information sharing for individualized
  treatment effects estimation.
\newblock In Dasgupta, S., Mandt, S., and Li, Y., editors, {\em Proceedings of
  The 27th International Conference on Artificial Intelligence and Statistics},
  volume 238 of {\em Proceedings of Machine Learning Research}, pages
  3529--3537. PMLR.

\bibitem[Chauhan et~al., 2024c]{chauhan2023brief}
Chauhan, V.~K., Zhou, J., Lu, P., Molaei, S., and Clifton, D.~A. (2024c).
\newblock A brief review of hypernetworks in deep learning.
\newblock {\em Artificial Intelligence Review}, 57(9):250.

\bibitem[Christiansen et~al., 2023]{christiansen2023device}
Christiansen, L.~B., Koch, S., Bauman, A., Toftager, M., Petersen, C.~B., and
  Schipperijn, J. (2023).
\newblock Device-based physical activity measures for population
  surveillance—issues of selection bias and reactivity.
\newblock {\em Frontiers in Sports and Active Living}, 5.

\bibitem[Christopher M.~Bishop, 2023]{bishop2024DL}
Christopher M.~Bishop, H.~B. (2023).
\newblock {\em Deep Learning: Foundations and Concepts}.
\newblock Springer Cham, Switzerland.

\bibitem[Cortes et~al., 2008]{cortes2008sample}
Cortes, C., Mohri, M., Riley, M., and Rostamizadeh, A. (2008).
\newblock Sample selection bias correction theory.
\newblock In {\em International conference on algorithmic learning theory},
  pages 38--53. Springer.

\bibitem[de~Jong et~al., 2021]{Jong2021appraising}
de~Jong, Y., Ramspek, C.~L., Zoccali, C., Jager, K.~J., Dekker, F.~W., and van
  Diepen, M. (2021).
\newblock Appraising prediction research: a guide and meta-review on bias and
  applicability assessment using the prediction model risk of bias assessment
  tool (probast).
\newblock {\em Nephrology}, 26(12):939--947.

\bibitem[de~Mathelin et~al., 2021]{de2021adapt}
de~Mathelin, A., Deheeger, F., Richard, G., Mougeot, M., and Vayatis, N.
  (2021).
\newblock Adapt: Awesome domain adaptation python toolbox.
\newblock {\em arXiv preprint arXiv:2107.03049}.

\bibitem[Dost, 2022]{dost2022selection}
Dost, K. (2022).
\newblock {\em Selection Bias Identification and Mitigation With No Ground
  Truth Information}.
\newblock PhD thesis, ResearchSpace@ Auckland.

\bibitem[Du et~al., 2022]{Du2022fair}
Du, W., Wu, X., and Tong, H. (2022).
\newblock Fair regression under sample selection bias.
\newblock In {\em IEEE International Conference on Big Data}, pages 1435--1444.

\bibitem[Elkan, 2001]{elkan2001foundations}
Elkan, C. (2001).
\newblock The foundations of cost-sensitive learning.
\newblock In {\em International joint conference on artificial intelligence},
  volume 17(1), pages 973--978. Lawrence Erlbaum Associates Ltd.

\bibitem[Ellen et~al., 2024]{Ellen2024}
Ellen, J.~G., Matos, J., Viola, M., Gallifant, J., Quion, J., {Anthony Celi},
  L., and {Abu Hussein}, N.~S. (2024).
\newblock Participant flow diagrams for health equity in ai.
\newblock {\em Journal of Biomedical Informatics}, page 104631.

\bibitem[Ganin et~al., 2016]{ganin2016domain}
Ganin, Y., Ustinova, E., Ajakan, H., Germain, P., Larochelle, H., Laviolette,
  F., Marchand, M., and Lempitsky, V. (2016).
\newblock Domain-adversarial training of neural networks.
\newblock {\em The journal of machine learning research}, 17(1):2096--2030.

\bibitem[Goetz et~al., 2024]{goetz2024generalization}
Goetz, L., Seedat, N., Vandersluis, R., and van~der Schaar, M. (2024).
\newblock Generalization—a key challenge for responsible ai in patient-facing
  clinical applications.
\newblock {\em npj Digital Medicine}, 7(1):126.

\bibitem[Greenland et~al., 1999]{greenland1999causal}
Greenland, S., Pearl, J., and Robins, J.~M. (1999).
\newblock Causal diagrams for epidemiologic research.
\newblock {\em Epidemiology}, pages 37--48.

\bibitem[Gronau, 1974]{gronau1974wage}
Gronau, R. (1974).
\newblock Wage comparisons--a selectivity bias.
\newblock {\em Journal of political Economy}, 82(6):1119--1143.

\bibitem[Guo et~al., 2017]{guo2017calibration}
Guo, C., Pleiss, G., Sun, Y., and Weinberger, K.~Q. (2017).
\newblock On calibration of modern neural networks.
\newblock In {\em International conference on machine learning}, pages
  1321--1330. PMLR.

\bibitem[Heckman, 1990]{heckman1990varieties}
Heckman, J. (1990).
\newblock Varieties of selection bias.
\newblock {\em The American Economic Review}, 80(2):313--318.

\bibitem[Heckman, 1976]{heckman1976common}
Heckman, J.~J. (1976).
\newblock The common structure of statistical models of truncation, sample
  selection and limited dependent variables and a simple estimator for such
  models.
\newblock In {\em Annals of economic and social measurement, volume 5, number
  4}, pages 475--492. NBER.

\bibitem[Heckman, 1979]{heckman1979sample}
Heckman, J.~J. (1979).
\newblock Sample selection bias as a specification error.
\newblock {\em Econometrica: Journal of the econometric society}, pages
  153--161.

\bibitem[Hendrickx et~al., 2024]{hendrickx2024machine}
Hendrickx, K., Perini, L., Van~der Plas, D., Meert, W., and Davis, J. (2024).
\newblock Machine learning with a reject option: A survey.
\newblock {\em Machine Learning}, pages 1--38.

\bibitem[Hernan and Robins, 2023]{hernan2023causal}
Hernan, M. and Robins, J. (2023).
\newblock {\em Causal Inference: What If}.
\newblock Chapman \& Hall/CRC monographs on statistics \& applied probability.
  Taylor \& Francis.

\bibitem[Hern{\'a}n et~al., 2004]{hernan2004structural}
Hern{\'a}n, M.~A., Hern{\'a}ndez-D{\'\i}az, S., and Robins, J.~M. (2004).
\newblock A structural approach to selection bias.
\newblock {\em Epidemiology}, pages 615--625.

\bibitem[Hong et~al., 2024]{Hong2024}
Hong, C., Liu, M., Wojdyla, D.~M., Hickey, J., Pencina, M., and Henao, R.
  (2024).
\newblock Trans-balance: Reducing demographic disparity for prediction models
  in the presence of class imbalance.
\newblock {\em Journal of Biomedical Informatics}, 149:104532.

\bibitem[Hosny et~al., 2018]{hosny2018artificial}
Hosny, A., Parmar, C., Quackenbush, J., Schwartz, L.~H., and Aerts, H.~J.
  (2018).
\newblock Artificial intelligence in radiology.
\newblock {\em Nature Reviews Cancer}, 18(8):500--510.

\bibitem[Huang et~al., 2006]{huang2006correcting}
Huang, J., Gretton, A., Borgwardt, K., Sch{\"o}lkopf, B., and Smola, A. (2006).
\newblock Correcting sample selection bias by unlabeled data.
\newblock {\em Advances in neural information processing systems}, 19.

\bibitem[J~Banasik and Thomas, 2003]{Banasik2003}
J~Banasik, J.~C. and Thomas, L. (2003).
\newblock Sample selection bias in credit scoring models.
\newblock {\em Journal of the Operational Research Society}, 54(8):822--832.

\bibitem[Jumper et~al., 2021]{jumper2021highly}
Jumper, J., Evans, R., Pritzel, A., Green, T., Figurnov, M., Ronneberger, O.,
  Tunyasuvunakool, K., Bates, R., {\v{Z}}{\'\i}dek, A., Potapenko, A., et~al.
  (2021).
\newblock Highly accurate protein structure prediction with alphafold.
\newblock {\em Nature}, 596(7873):583--589.

\bibitem[Kingma and Ba, 2014]{kingma2014adam}
Kingma, D.~P. and Ba, J. (2014).
\newblock Adam: A method for stochastic optimization.
\newblock {\em arXiv preprint arXiv:1412.6980}.

\bibitem[Kompa et~al., 2021]{kompa2021second}
Kompa, B., Snoek, J., and Beam, A.~L. (2021).
\newblock Second opinion needed: communicating uncertainty in medical machine
  learning.
\newblock {\em NPJ Digital Medicine}, 4(1):4.

\bibitem[Kouw and Loog, 2019]{kouw2019review}
Kouw, W.~M. and Loog, M. (2019).
\newblock A review of domain adaptation without target labels.
\newblock {\em IEEE transactions on pattern analysis and machine intelligence},
  43(3):766--785.

\bibitem[Kundu et~al., 2023]{kundu2023framework}
Kundu, R., Shi, X., Morrison, J., and Mukherjee, B. (2023).
\newblock A framework for understanding selection bias in real-world healthcare
  data.
\newblock {\em arXiv preprint arXiv:2304.04652}.

\bibitem[Liu et~al., 2023]{liu2023combining}
Liu, X., Morelli, D., Littlejohns, T.~J., Clifton, D.~A., and Clifton, L.
  (2023).
\newblock Combining machine learning with cox models to identify predictors for
  incident post-menopausal breast cancer in the uk biobank.
\newblock {\em Scientific Reports}, 13(1):9221.

\bibitem[Long and Ha, 2022]{long2022sample}
Long, J.~P. and Ha, M.~J. (2022).
\newblock Sample selection bias in evaluation of prediction performance of
  causal models.
\newblock {\em Statistical Analysis and Data Mining: The ASA Data Science
  Journal}, 15(1):5--14.

\bibitem[Mei and Xia, 2019]{mei2019knowledge}
Mei, J. and Xia, E. (2019).
\newblock Knowledge learning symbiosis for developing risk prediction models
  from regional ehr repositories.
\newblock In {\em MEDINFO 2019: Health and Wellbeing e-Networks for All}, pages
  258--262. IOS Press.

\bibitem[Minderer et~al., 2021]{minderer2021revisiting}
Minderer, M., Djolonga, J., Romijnders, R., Hubis, F., Zhai, X., Houlsby, N.,
  Tran, D., and Lucic, M. (2021).
\newblock Revisiting the calibration of modern neural networks.
\newblock {\em Advances in Neural Information Processing Systems},
  34:15682--15694.

\bibitem[Navarro et~al., 2021]{Navarron2021}
Navarro, C. L.~A., Damen, J. A.~A., Takada, T., Nijman, S. W.~J., Dhiman, P.,
  Ma, J., Collins, G.~S., Bajpai, R., Riley, R.~D., Moons, K. G.~M., and Hooft,
  L. (2021).
\newblock Risk of bias in studies on prediction models developed using
  supervised machine learning techniques: systematic review.
\newblock {\em BMJ}, 375.

\bibitem[Parsons et~al., 2023]{parsons2023independent}
Parsons, R.~E., Liu, X., Collister, J.~A., Clifton, D.~A., Cairns, B.~J., and
  Clifton, L. (2023).
\newblock Independent external validation of the qrisk3 cardiovascular disease
  risk prediction model using uk biobank.
\newblock {\em Heart}, 109(22):1690--1697.

\bibitem[Paszke et~al., 2019]{paszke2019pytorch}
Paszke, A., Gross, S., Massa, F., Lerer, A., Bradbury, J., Chanan, G., Killeen,
  T., Lin, Z., Gimelshein, N., Antiga, L., et~al. (2019).
\newblock Pytorch: An imperative style, high-performance deep learning library.
\newblock {\em Advances in neural information processing systems}, 32.

\bibitem[Pearl and Mackenzie, 2018]{pearl2018book}
Pearl, J. and Mackenzie, D. (2018).
\newblock {\em The book of why: the new science of cause and effect}.
\newblock Basic books.

\bibitem[Porta, 2014]{porta2014dictionary}
Porta, M. (2014).
\newblock {\em A dictionary of epidemiology}.
\newblock Oxford university press.

\bibitem[Rajkomar et~al., 2019]{rajkomar2019machine}
Rajkomar, A., Dean, J., and Kohane, I. (2019).
\newblock Machine learning in medicine.
\newblock {\em New England Journal of Medicine}, 380(14):1347--1358.

\bibitem[Rojas-Saunero et~al., 2023]{rojas2023selection}
Rojas-Saunero, L.~P., Glymour, M.~M., and Mayeda, E.~R. (2023).
\newblock Selection bias in health research: Quantifying, eliminating, or
  exacerbating health disparities?
\newblock {\em Current Epidemiology Reports}, pages 1--10.

\bibitem[Salaun et~al., 2023]{salaun2023interpretable}
Salaun, A., Knight, S., Wingfield, L.~R., and Zhu, T. (2023).
\newblock Interpretable machine learning in kidney offering: Multiple outcome
  prediction for accepted offers.
\newblock {\em medRxiv}, pages 2023--08.

\bibitem[Shimodaira, 2000]{shimodaira2000improving}
Shimodaira, H. (2000).
\newblock Improving predictive inference under covariate shift by weighting the
  log-likelihood function.
\newblock {\em Journal of statistical planning and inference}, 90(2):227--244.

\bibitem[Smith, 2020]{smith2020selection}
Smith, L.~H. (2020).
\newblock Selection mechanisms and their consequences: understanding and
  addressing selection bias.
\newblock {\em Current Epidemiology Reports}, 7:179--189.

\bibitem[Stekhoven and Buehlmann, 2012]{Daniel2012MissForest}
Stekhoven, D.~J. and Buehlmann, P. (2012).
\newblock Missforest - non-parametric missing value imputation for mixed-type
  data.
\newblock {\em Bioinformatics}, 28(1):112--118.

\bibitem[Sugiyama et~al., 2007]{sugiyama2007direct}
Sugiyama, M., Nakajima, S., Kashima, H., Buenau, P., and Kawanabe, M. (2007).
\newblock Direct importance estimation with model selection and its application
  to covariate shift adaptation.
\newblock {\em Advances in neural information processing systems}, 20.

\bibitem[Swanson, 2012]{swanson2012uk}
Swanson, J.~M. (2012).
\newblock The uk biobank and selection bias.
\newblock {\em The Lancet}, 380(9837):110.

\bibitem[Thakur et~al., 2018]{thakur2018use}
Thakur, S.~S., Li, H., Chan, A.~M., Tudor, R., Bigras, G., Morris, D., Enwere,
  E.~K., and Yang, H. (2018).
\newblock The use of automated ki67 analysis to predict oncotype dx
  risk-of-recurrence categories in early-stage breast cancer.
\newblock {\em PLoS One}, 13(1):e0188983.

\bibitem[Topol, 2019]{topol2019high}
Topol, E.~J. (2019).
\newblock High-performance medicine: the convergence of human and artificial
  intelligence.
\newblock {\em Nature medicine}, 25(1):44--56.

\bibitem[Vandersluis and Savulescu, 2024]{vandersluis2024selective}
Vandersluis, R. and Savulescu, J. (2024).
\newblock The selective deployment of ai in healthcare: An ethical algorithm
  for algorithms.
\newblock {\em Bioethics}, 38(5):391--400.

\bibitem[Vella, 1998]{vella1998estimating}
Vella, F. (1998).
\newblock Estimating models with sample selection bias: a survey.
\newblock {\em Journal of Human Resources}, pages 127--169.

\bibitem[Vogel et~al., 2020]{vogel2020weighted}
Vogel, R., Achab, M., Clémençon, S., and Tillier, C. (2020).
\newblock Weighted empirical risk minimization: Sample selection bias
  correction based on importance sampling.

\bibitem[Vokinger et~al., 2021]{vokinger2021mitigating}
Vokinger, K.~N., Feuerriegel, S., and Kesselheim, A.~S. (2021).
\newblock Mitigating bias in machine learning for medicine.
\newblock {\em Communications medicine}, 1(1):25.

\bibitem[Wagenaar et~al., 2021]{wagenaar2021improved}
Wagenaar, D., Hermawan, T., van~den Homberg, M.~J., Aerts, J.~C., Kreibich, H.,
  de~Moel, H., and Bouwer, L.~M. (2021).
\newblock Improved transferability of data-driven damage models through sample
  selection bias correction.
\newblock {\em Risk analysis}, 41(1):37--55.

\bibitem[Weuve et~al., 2012]{weuve2012accounting}
Weuve, J., Tchetgen, E. J.~T., Glymour, M.~M., Beck, T.~L., Aggarwal, N.~T.,
  Wilson, R.~S., Evans, D.~A., and de~Leon, C. F.~M. (2012).
\newblock Accounting for bias due to selective attrition: the example of
  smoking and cognitive decline.
\newblock {\em Epidemiology (Cambridge, Mass.)}, 23(1):119.

\bibitem[Wolff et~al., 2019]{wolff2019probast}
Wolff, R.~F., Moons, K.~G., Riley, R.~D., Whiting, P.~F., Westwood, M.,
  Collins, G.~S., Reitsma, J.~B., Kleijnen, J., Mallett, S., and Group†, P.
  (2019).
\newblock Probast: a tool to assess the risk of bias and applicability of
  prediction model studies.
\newblock {\em Annals of internal medicine}, 170(1):51--58.

\bibitem[Yu and Eng, 2020]{yu2020one}
Yu, A.~C. and Eng, J. (2020).
\newblock One algorithm may not fit all: how selection bias affects machine
  learning performance.
\newblock {\em Radiographics}, 40(7):1932--1937.

\bibitem[Zadrozny, 2004]{zadrozny2004learning}
Zadrozny, B. (2004).
\newblock Learning and evaluating classifiers under sample selection bias.
\newblock In {\em Proceedings of the twenty-first international conference on
  Machine learning}, page 114.

\bibitem[Zhelonkin et~al., 2016]{zhelonkin2016robust}
Zhelonkin, M., Genton, M.~G., and Ronchetti, E. (2016).
\newblock Robust inference in sample selection models.
\newblock {\em Journal of the Royal Statistical Society Series B: Statistical
  Methodology}, 78(4):805--827.

\end{thebibliography}

\end{document}